\documentclass{article}

\usepackage{graphicx}
\usepackage{subfigure}

\usepackage{amsmath,amsfonts,bm}

\def\eqref#1{equation~\ref{#1}}
\def\1{\bm{1}}

\def\rvepsilon{{\mathbf{\epsilon}}}

\def\rva{{\mathbf{a}}}
\def\rvb{{\mathbf{b}}}

\def\rvp{{\mathbf{p}}}
\def\rvq{{\mathbf{q}}}

\def\rvs{{\mathbf{s}}}

\def\rvw{{\mathbf{w}}}
\def\rvx{{\mathbf{x}}}

\def\vmu{{\bm{\mu}}}
\def\vtheta{{\bm{\theta}}}

\def\vf{{\bm{f}}}

\def\vh{{\bm{h}}}

\def\vl{{\bm{l}}}

\def\mA{{\bm{A}}}
\def\mB{{\bm{B}}}

\def\mF{{\bm{F}}}
\def\mG{{\bm{G}}}

\def\mI{{\bm{I}}}
\def\mJ{{\bm{J}}}

\def\mL{{\bm{L}}}
\def\mM{{\bm{M}}}

\def\mP{{\bm{P}}}

\def\mSigma{{\bm{\Sigma}}}

\DeclareMathAlphabet{\mathsfit}{\encodingdefault}{\sfdefault}{m}{sl}
\SetMathAlphabet{\mathsfit}{bold}{\encodingdefault}{\sfdefault}{bx}{n}

\def\gL{{\mathcal{L}}}

\def\gN{{\mathcal{N}}}

\def\gU{{\mathcal{U}}}

\newcommand{\E}{\mathbb{E}}

\newcommand{\R}{\mathbb{R}}

\newcommand{\KL}{D_{\mathrm{KL}}}

\usepackage[utf8]{inputenc} % allow utf-8 input
\usepackage[T1]{fontenc}    % use 8-bit T1 fonts
\usepackage{hyperref}       % hyperlinks
\usepackage{url}            % simple URL typesetting
\usepackage{booktabs}       % professional-quality tables
\usepackage{amsfonts}       % blackboard math symbols
\usepackage{nicefrac}       % compact symbols for 1/2, etc.
\usepackage{microtype}      % microtypography
\usepackage{xcolor}         % colors

\usepackage{amsmath}
\usepackage{amssymb}
\usepackage{mathtools}
\usepackage{amsthm}

\usepackage{physics}
\usepackage{booktabs}
\usepackage{wrapfig}
\usepackage{pifont}
\usepackage{caption}
\usepackage{lipsum}
\usepackage{multirow}
\usepackage{algorithm}

\usepackage[top=2.5cm,bottom=2.3cm,inner=2.3cm,outer=2.3cm,bindingoffset=0.5cm,footskip=0.55cm]{geometry}
\theoremstyle{plain}

\theoremstyle{definition}

\theoremstyle{remark}

\newtheorem{lem}{Lemma}

\DeclarePairedDelimiterX{\infdivx}[2]{(}{)}{%
  #1\;\delimsize\|\;#2%
}
\newcommand{\infdiv}{\KL\infdivx}

\title{Generative Modelling with High-Order Langevin Dynamics}

\author{%
  Ziqiang Shi, Rujie Liu\\
  FRDC\\
 Beijing, China \\
  \texttt{shiziqiang@fujitsu.com} \\
}

\date{}

\begin{document}

\maketitle

\begin{abstract}
  Diffusion generative modelling (DGM) based on stochastic 
  differential equations (SDEs) with 
  score matching has achieved unprecedented results in data 
  generation.
   In this paper, we propose a novel fast high-quality 
   generative modelling method 
   based on high-order
   Langevin dynamics (HOLD) with score matching. 
   This motive is proved by third-order
   Langevin dynamics. By augmenting the 
   previous SDEs, e.g.
   variance exploding or variance preserving SDEs 
   for single-data variable processes, HOLD can simultaneously 
   model position, velocity, and 
   acceleration, thereby improving the quality 
   and speed of the data 
   generation at the same time. 
   HOLD is composed of one Ornstein-Uhlenbeck process 
   and two Hamiltonians, 
   which reduce the mixing time by two orders of magnitude. 
   Empirical experiments for unconditional image generation on the
 public data set CIFAR-10 and CelebA-HQ show that the effect is significant in
 both Frechet inception distance (FID) and negative log-likelihood, 
 and achieves the
 state-of-the-art FID of 1.85 on CIFAR-10. 
\end{abstract}

\section{Introduction}
\label{sec:intro}

Diffusion generative model (DGM) is characterized by gradually
adding noise to the target data until it becomes
the data conforming to some simple prior distribution.
The earliest such model is
proposed by~\cite{sohl2015deep} in the field of image generation, 
and then generalized 
by~\cite{ho2020denoising} and~\cite{song2020score}, 
which proposed
DGM based on denoising score matching with 
Langevin dynamics to 
achieve the quality that 
surpasses generative adversarial 
network (GAN)~\cite{goodfellow2014generative} or 
variational autoencoder (VAE)~\cite{kingma2014auto}-based 
generation methods in the field
of high-resolution high-quality image generation.
Then, in just two years, DGM made unprecedented progress 
and has achieved success in many fields such as
image %generation~\cite{ho2020denoising} and 
editing~\cite{meng2021sdedit}, % molecular 
% modelling~\cite{xu2022geodiff}, speech 
% generation~\cite{wu2022itowave}, music 
% synthesis~\cite{liu2022diffsinger},
% video modelling~\cite{ho2022video}, 
natural language 
processing~\cite{li2022diffusion}, 
text-to-image~\cite{rombach2022high}, 
and so on.

In addition to applying DGM to different fields, another aspect 
of research is how to improve the existing DGM itself, so that 
the quality of the generated data is higher and the speed of 
generation is faster. 
There are two types of DGM, one is 
discrete~\cite{ho2020denoising,bao2022estimating}
and the other is continuous~\cite{song2020score,dockhorn2021score}.
For an efficient discrete DGM, we need to design an ingenious 
noise scheduling with a finite length Markov chain to gradually 
realize the mutual conversion between data and white noise.
While for continuous DGM, the main task is to design a good stochastic 
differential equation (SDE) and a corresponding integrator. 
This paper mainly explores new possibilities in continuous DGM.

\begin{figure}[th]%[htp]
  %\vspace{-0.4in}
  \centering
  %\begin{center}
  \hspace{-5mm}
  \includegraphics[width=1.0\linewidth]{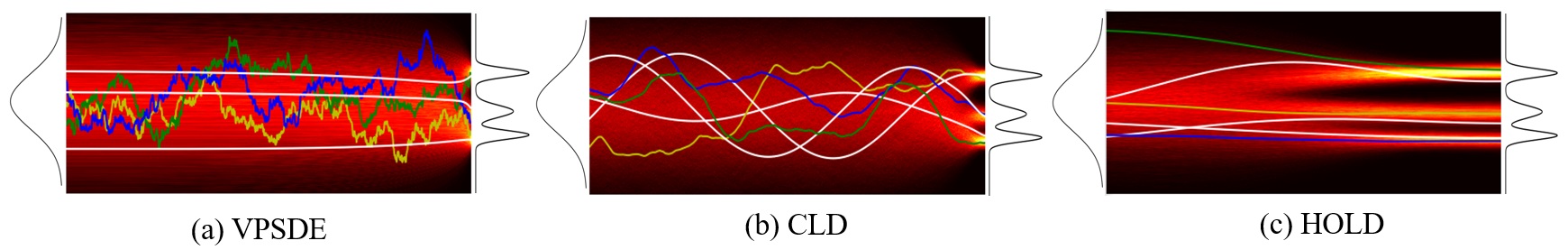}
  \hspace{-5mm}
  %\end{center}
  %\vspace{-2.5mm}
  \caption{
  Comparison of solution paths between HOLD-based DGM and other 
  methods on 1D data. 
  The paths in white are obtained by integrating the corresponding 
  ODE flows. 
  The coloured paths are obtained by solving the corresponding 
  backward generative SDE with Euler-Maruyama method.
  }
  \label{fig:1d_diffusion_comparision}
  %\vspace{-2.5mm}
  \end{figure}

Recently there has been a lot of research work on continuous 
DGM, such as designing different integrators, or 
fast solvers for revserse-time SDEs with trained scoring 
models~\cite{dockhorn2022genie,liu2022pseudo,bao2022estimating,lu2022dpm}.
This paper focuses on another direction, which is to improve 
the SDE itself in the forward diffusion 
process~\cite{dockhorn2021score,wang2021deep}.
Inspired by critically-damped Langevin 
diffusion (CLD)~\cite{dockhorn2021score} and the work of~\cite{mou2021high}, we propose
high-order Langevin dynamics (HOLD)
to further decouple the position variables from the 
Brownian motion by introducing new intermediate variables.
Most of the SDEs used for generate modelling are first-order 
for data variables (such as images or 3D models, etc.),
but theoretically, we can formally introduce any number of 
intermediate variables into these SDEs, such as
\begin{equation} 
  \begin{cases}
    &d\rvx_1  = \rvx_2 dt, \\
    &d\rvx_2 =  -L^{-1}\nabla V(\rvx_1)  dt  +  \gamma\rvx_3 dt, \\
    %&d\rvx_3 =  -L^{-1}\nabla V(\rvx_2)  dt  +  \gamma\rvx_4dt, \\
    &\quad \cdots \\
    &d\rvx_{n-1} =  -L^{-1}\nabla V(\rvx_{n-2})  dt  +  \gamma\rvx_n dt, \\
    &d\rvx_n =  -\gamma\rvx_{n-1}dt - \xi\rvx_n dt  + \sqrt{2\xi L^{-1}} d\rvw_t,
  \end{cases}\label{eq:nth_order_langevin}
  \end{equation}
so that the Brownian motion $\sqrt{2\xi L^{-1}} d\rvw_t$ can 
only affect the position (data)
variable $\rvx_1$ after $n-1$ times of conduction
through the intermediate variables $\rvx_2,\cdots,\rvx_n$, then 
the final path 
of the 
position variable will be much smoother.
Eq.~(\ref{eq:nth_order_langevin}) is called high-order
Langevin dynamics (HOLD), which will be used as the forward diffusion
dynamics for our newly proposed DGM.
In HOLD, two adjacent variables can form a Hamiltonian dynamics  
%
%\vspace{0.2cm}
\begin{equation} 
\vspace{0.2cm}
  \begin{cases}
    &d\rvx_{i-1} =  \gamma\rvx_i dt, \\
    &d\rvx_i =  -L^{-1}\nabla V(\rvx_{i-1})  dt 
    \end{cases}\label{eq:ith_hamiltonian}
\vspace{-0.2cm}
  \end{equation}
and the 
Brownian motion only directly affects the last variable $\rvx_n$. 
That is to say, 
HOLD consists of $n-1$ Hamiltonian components and one 
Ornstein-Uhlenbeck (OU) process. 
And each of these Hamiltonian dynamics can quickly traverse the 
data space and reduce 
the mixing time of the system~\cite{neal2011mcmc}.
To verify our idea, we use a third-order Langevin 
dynamics~\cite{mou2021high} (refer 
to Eq.~(\ref{eq:third_order_langevin}) and for 
convenience it is also 
referred to as 
  HOLD hereinafter
  %, and the order is no longer distinguished
  ) to 
  implement the forward diffusion process. 
This HOLD can simulate position, velocity, and acceleration 
simultaneously, where the position variable does not directly 
depend on the energy gradient and Brownian component, 
which are separated by velocity and acceleration. 
Only the acceleration depends explicitly on the random component.
Figure~\ref{fig:1d_diffusion_comparision}  gives a one-dimensional 
example, and it can be seen 
that the solution path of DGM based on HOLD is gentler and 
smoother than that of CLD and variance preservering (VP) SDE.
This results in fewer 
steps to generate high-quality data, which means faster generation. 

Our HOLD has been extensively testified on sevaral public 
benchmarks. It is shown that HOLD-based DGM admits easy-to-tune 
hyperparameters,
better synthesis quality, shorter mixing times, and faster 
sampling. 
Our tailor-made solver for HOLD-DGM far surpasses other methods.
On the CIFAR-10 and CelebA real-world image modelling task, 
HOLD-based DGM outperforms various state-of-the-art methods in 
synthesis quality under similar computation resources.
Last but not least, we find an excellent computationally 
efficient instance for HOLD-DGM in the vast ocean of third-order 
stochastic Langevin dynamics.

\section{Background}
\label{sec:background}

It\^o linear SDE 
is an excellent and tractable model for 
converting between data of different probability 
distributions~\cite{song2020score}. 
The general form of It\^o linear SDE  is as follows
\begin{align}\label{eq:diffusion_sde}
  d\rvq_t &= \vf(t)\rvq_t \, dt + \mG(t) \, d\rvw_t, \quad t \in [0, T],
\end{align}
where  $\rvq_t\in  \R^{d}$ is the position vector of a particle 
(represents the data that needs to be generated in DGM, 
such as images or 3D models, etc.), 
$\vf(t), \mG(t) \in \R^{d\times d}$,
$\vf(t)\rvq_t$ is the drift coefficient,
 $\mG(t)$ is the diffusion coefficient,
 and $\rvw_t$ is the standard Wiener process. 
 First-order continuous-time Langevin dynamics 
 are a special case of 
 stochastic process represented by the 
SDE~(\ref{eq:diffusion_sde})~\cite{mou2021high}.
Let $p(\rvq_t, t)$ be the probability density of the 
stochastic process $\rvq_t$ at time $t$. 
This SDE~(\ref{eq:diffusion_sde}) transforms the beginning 
data distribution 
$p(\rvq_0, 0)$ into a final simple tractable latent 
distribution 
$p(\rvq_T, T)$
by gradually adding the noise from $\rvw_t$.

  \vspace{-0.2cm}

If the solution process $\rvq_t$ can be reversed in time, the target 
image or 3D data can be generated from a simple 
latent distribution. 
Indeed the time-reverse process is the solution of the 
corresponding backward SDE~\cite{anderson1982reverse}
\begin{align}
%  &d\rvq_t=\left[\vf(t)\rvq_t-\mG(t)\mG(t)^\top\nabla_{\rvq_t}\log p(\rvq_t, t)\right]dt \nonumber
%  \\&+\mG(t)d\overline{\rvw}_t 
 d\rvq_t=\left[\vf(t)\rvq_t-\mG(t)\mG(t)^\top\nabla_{\rvq_t}\log p(\rvq_t, t)\right]dt 
 +\mG(t)d\overline{\rvw}_t 
 \label{eq:reverse_diffusion_sde}
\end{align}
for $0\leq t\leq T$,
where 
$\overline{\rvw}_t$ is the standard Wiener process in reverse time.
Therefore, it can be seen that the key to generative 
modelling with SDE lies in the calculations of 
$\nabla_{\rvq_t}\log p(\rvq_t, t)$, $(0\leq t\leq T)$, which is always 
called score function~\cite{hyvarinen2005estimation,song2020score}. 
And the score function can be obtained by optimizing the denoising score matching (DSM) 
loss~\cite{hyvarinen2005estimation,vincent2011connection}
\begin{align}
  % \text{DSM loss}= \mathbb{E}_{t\sim [0,T]}\mathbb{E}_{\rvq_0\sim p_{data}(\rvq_0, 0)}  
  % \mathbb{E}_{\rvq_t\sim p(\rvq_t|\rvq_0)} \nonumber \\
  %  \left[ \frac{1}{2}\parallel \mathfrak{S}_{\theta}(\rvq_t,t) -\nabla_{\rvq_t}  \log  p(\rvq_t|\rvq_0)\parallel^2 \right].
   \text{DSM loss}= \mathbb{E}_{t\sim [0,T]}\mathbb{E}_{\rvq_0\sim p_{data}(\rvq_0, 0)}  
   \mathbb{E}_{\rvq_t\sim p(\rvq_t|\rvq_0)} 
    \left[ \frac{1}{2}\parallel \mathfrak{S}_{\theta}(\rvq_t,t) -\nabla_{\rvq_t}  \log  p(\rvq_t|\rvq_0)\parallel^2 \right]. 
   \label{eq:dsm_loss}
\end{align}
After the score function $\mathfrak{S}_{\theta}(\rvq_t,t)$ is learned,
then the Langevin Markov chain Monte Carlo (MCMC) 
% \begin{align}
%  \rvq_{t+} = \rvq_{t}+\eta_t\mathfrak{S}_{\theta}(\rvq_t,t)+\sqrt{2\eta_t}\rvepsilon_t
%   \label{eq:first_order_langevin}
% \end{align}
or the discretized time-reverse 
SDE~(\ref{eq:reverse_diffusion_sde}) can be used to 
generate large-scale data in target distribution. 
% Here 
% $\rvepsilon_{t} \sim \gN(\bm{0}, \mI_{d})$ and 
% $\lim_{t\rightarrow\infty} \eta_t= 0$. 

\section{Generative Modelling with High-Order Langevin Dynamics}

First-order continuous-time Langevin dynamics and score matching  
are key to the success of SDE-based DGM~\cite{song2020score}
and its following work~\cite{song2021maximum}. 
Although the continuous-time Langevin dynamics converges to the 
target distribution at an exponential rate, the error caused by 
discretization in practice makes the mixing time only 
%$O (\frac{d}{\epsilon^2})$~\cite{mou2021high},
$O (d/\epsilon^2)$~\cite{mou2021high},
where $\epsilon > 0$ is the Wasserstein distance 
from the target distribution.
We propose to use high-order lifting to 
generalize and move this idea to higher-order continuous dynamics, 
such that momentum (also referred to
as velocity) and acceleration are introduced to accelerate 
the convergence and 
enhance the sampling quality of position vectors.

\subsection{Third-Order Langevin Dynamics}
\label{sec:told}

We model the forward diffusion process as the solution of the 
following 
third-order It\^o SDE 
\begin{equation} 
\begin{cases}
  d\rvq_t & = \rvp_t dt, \\
  d\rvp_t & =  -L^{-1}\nabla U(\rvq_t)  dt  +  \gamma\rvs_t dt, \\
  d\rvs_t & =  -\gamma\rvp_tdt - \xi\rvs_t dt  + \sqrt{2\xi L^{-1}} d\rvw_t
\end{cases}\label{eq:third_order_langevin}
\end{equation}
where $\rvq_t, \rvp_t, \rvs_t\in \R^{d}$ are the position, 
momentum 
(first row in Eq.~(\ref{eq:third_order_langevin}) implies that 
 $\rvp_t$ is the time derivative of position $\rvq_t$),
and acceleration processes respectively, $L$ is the 
Lipschitz constant of 
$U$, $\gamma>0$ is the friction coefficient, and 
$\xi>0$ is an algorithmic parameter. 
In the modelling equation of
momentum,
$U(\rvq_t): \R^{d} \to \R$ is the potential function related 
to the 
position vector. In this paper, we 
employ $U(\rvq_t)=\frac{L}{2}\norm{\rvq_t}$
which is strongly convex and Lipschitz smooth.
Let $\rvx_t = (\rvq_t, \rvp_t, \rvs_t)^\top \in \R^{3d}$, 
the SDE~(\ref{eq:third_order_langevin}) can be written
in matrix form
$d\rvx_t = \vf(t)\rvx_t \, dt + \mG(t) \, d\rvw_t$, 
where $\vf(t)$ and $\mG(t)$ are
\begin{align}
\begin{pmatrix} 0 &1&0\\ 
        -1 &0&\gamma\\ 
        0 & - \gamma & -\xi \end{pmatrix} \otimes \mI_d, 
\begin{pmatrix} 0 &0 &0\\ 
        0 &0 &0\\ 
        0  & 0 &\sqrt{2\xi L^{-1}} \end{pmatrix} \otimes \mI_d,
        \label{eq:hold_coefficients}
    \end{align}
% \begin{align}
%   &\vf(t) \coloneqq  \begin{pmatrix} 0 &1&0\\ 
%     -1 &0&\gamma\\ 
%     0 & - \gamma & -\xi \end{pmatrix} \otimes \mI_d, 
%     \label{eq:hold_drift_coefficients} \\
%   &\mG(t) \coloneqq \begin{pmatrix} 0 &0 &0\\ 
%     0 &0 &0\\ 
%     0  & 0 &\sqrt{2\xi L^{-1}} \end{pmatrix} \otimes \mI_d,
%     \label{eq:hold_diff_coefficients}
% \end{align}
%
respectively, here $\otimes$ denotes the Kronecker product.
Considering the computability, especially the 
complexity and the expressiveness, after trial 
and error, we let $\gamma^2=1+\frac{\xi^2}{4}$ in this paper. 
Since then $\vf(t)$ 
has three different real 
eigenvalues in elegant and simple closed form,  
which are $-\frac{\xi}{2}$ and $-\frac{\xi\pm \sqrt{\xi^2-32}}{4}$. 
This will make the mean and variance of the transition densities
\begin{align} 
% &p(\rvx_t|\rvx_0, t)=\gN(\rvx_t; \vmu_t, \mSigma_t), 
% \quad \mSigma_t = \Sigma_t  \otimes \mI_d, 
% \label{eq:transition_density}\\ &\Sigma_t  = 
% \begin{pmatrix} \Sigma_t^{qq} & \Sigma_t^{qp} & \Sigma_t^{qs}\\ 
%   \Sigma_t^{qp} & \Sigma_t^{pp} & \Sigma_t^{ps}\\ 
%   \Sigma_t^{qs} & \Sigma_t^{ps} & \Sigma_t^{ss}  \end{pmatrix}
&p(\rvx_t|\rvx_0, t)=\gN(\rvx_t; \vmu_t, \mSigma_t), 
\quad \mSigma_t = \Sigma_t  \otimes \mI_d, 
\label{eq:transition_density}\\ &\Sigma_t  = 
\begin{pmatrix} \Sigma_t^{qq} & \Sigma_t^{qp} & \Sigma_t^{qs}\\ 
  \Sigma_t^{qp} & \Sigma_t^{pp} & \Sigma_t^{ps}\\ 
  \Sigma_t^{qs} & \Sigma_t^{ps} & \Sigma_t^{ss}  \end{pmatrix}
\end{align}
of the solution 
process $\rvx_t$ feasible to calculate.
It can be shown that the prior distribution 
(or equilibrium distribution) of 
this diffusion is $p_{\infty }(\rvx)=\mathcal{N}(\rvq;\bm{0}_d,\frac{\xi L^{-1}}{6}\mI_d)\,\mathcal{N}(\rvp;\bm{0}_d,\frac{\xi L^{-1}}{6}\mI_d)\,\mathcal{N}(\rvs;\bm{0}_d,\frac{\xi L^{-1}}{6}\mI_d)$.
As introduced in Section~\ref{sec:intro}, the dynamics of 
Eq.~(\ref{eq:third_order_langevin}) 
is referred to as HOLD hereinafter.

There are two Hamiltonian components 
 $(\rvp_tdt, -\rvq_tdt)^\top$ and 
$(\gamma\rvs_tdt, -\gamma\rvp_tdt)^\top$, and
one Ornstein-Uhlenbeck (OU) process 
$(- \xi\rvs_t dt  + \sqrt{2\xi L^{-1}} d\rvw_t)$ in 
HOLD~(\ref{eq:third_order_langevin})
with coefficients~(\ref{eq:hold_coefficients}). 
Hamiltonian components can make two successive sampling points 
in the Markov chain far apart, 
so that a representative sampling of a canonical distribution 
can be obtained with fewer iterations.
Hamiltonian dynamics allows the Markov 
chain to explore target distribution much more 
efficiently~\cite{neal2011mcmc}.
Two Hamiltonian can lead to twice the convergence speedup.
The OU process has a bounded variance and 
admits a stationary probability distribution.
It has a tendency to move back 
towards a central location, with a greater attraction 
when the process is further away from the 
center~\cite{sarkka2019applied}. That is to say, all three 
components that make up HOLD contribute to the fast traversal 
of the solution space.

% \begin{wrapfigure}{r}{0.55\textwidth}
%   \vspace{-20pt}
%   \begin{center}
%   \includegraphics[width=0.5\textwidth]{hold_1d_diffusion_qps.png}
%   \end{center}
%   \vspace{-5pt}
%   \caption{In 1D case, the solution (data generation) paths of the 
%   position (white), velocity (yellow), and 
%   acceleration (green) variables in HOLD.
% }
% \label{fig:hold_1d_diffusion_qps}
%   \vspace{-10pt}
% \end{wrapfigure}

\begin{figure}[th]%[htp]
  %\vspace{-0.4in}
  \centering
  %\begin{center}
  \hspace{-5mm}
  \includegraphics[width=1.0\linewidth]{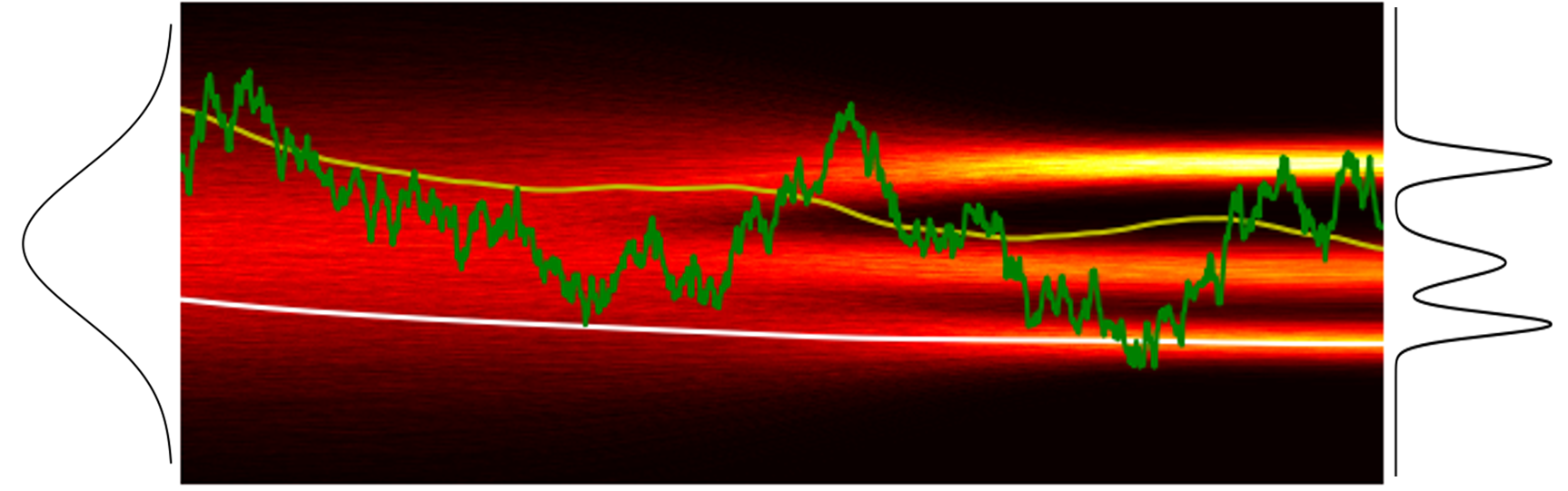}
  \hspace{-5mm}
  %\end{center}
  %\vspace{-2.5mm}
  \caption{
    In 1D case, the solution (data generation) paths of the 
    position (white), velocity (yellow), and 
    acceleration (green) variables in HOLD.
  }
  \label{fig:hold_1d_diffusion_qps}
  %\vspace{-2.5mm}
  \end{figure}
  \vspace{-0.2cm}

  \begin{figure}[!h]
    \centering
    \subfigure[2D multi-Swiss rolls generated by CLD-based DGM 
    at 6 moments along the solution path.]{
    \begin{minipage}[t]{1.0\linewidth}
    \centering
    \includegraphics[width=5.0in]{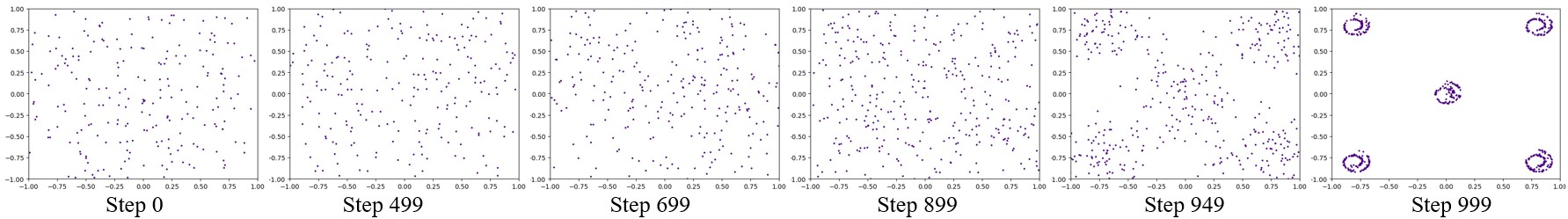}
    \label{fig:cld_2d_swiss_paths}
    \end{minipage}%
    }%
  
  \subfigure[2D multi-Swiss rolls generated by HOLD-based DGM 
  at 6 moments along the solution path.]{
    \begin{minipage}[t]{1.0\linewidth}
    \centering
    \includegraphics[width=5.0in]{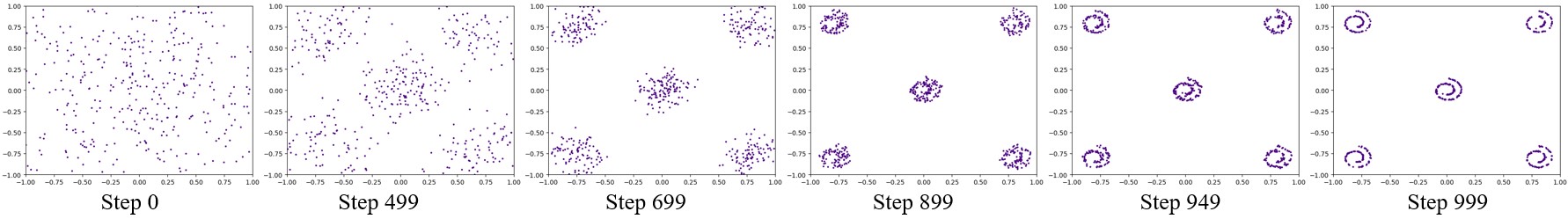}
    \label{fig:hold_2d_swiss_paths}
    \end{minipage}%
    }%
    \centering
    \caption{Comparison of different moments (number of steps) 
    on the solution path 
    generated by CLD and HOLD-based DGM on 2D multi-Swiss rolls.}
   \label{fig:2d_swiss_paths}
    \end{figure}

\begin{figure}[!h]
  \centering
  \subfigure[Data]{
  \begin{minipage}[t]{0.5\linewidth}
  \centering
  \includegraphics[width=1.6in]{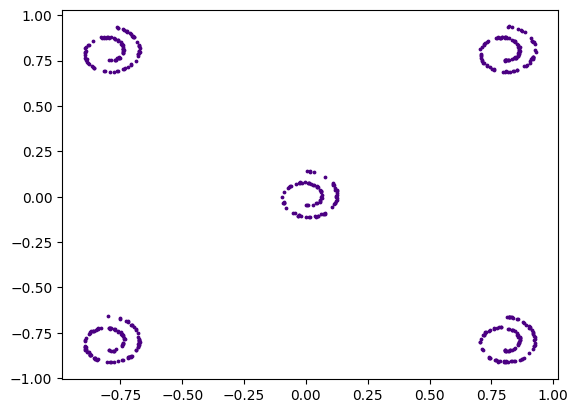}
  \label{fig:data_2d_swiss_sample}
  \end{minipage}%
  }%
  \subfigure[VP]{
  \begin{minipage}[t]{0.5\linewidth}
  \centering
  \includegraphics[width=1.6in]{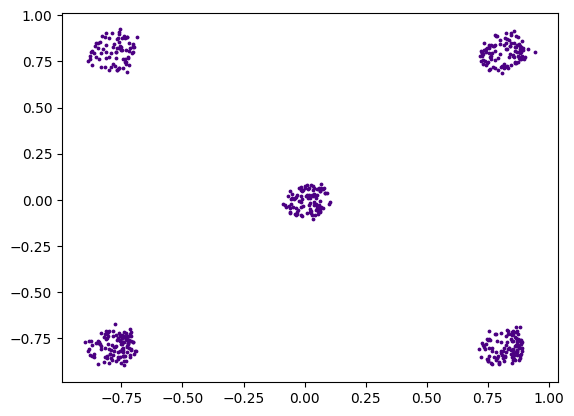}
  \label{fig:vp_2d_swiss_sample}
  \end{minipage}%
  }%

  \subfigure[CLD]{
    \begin{minipage}[t]{0.5\linewidth}
    \centering
    \includegraphics[width=1.6in]{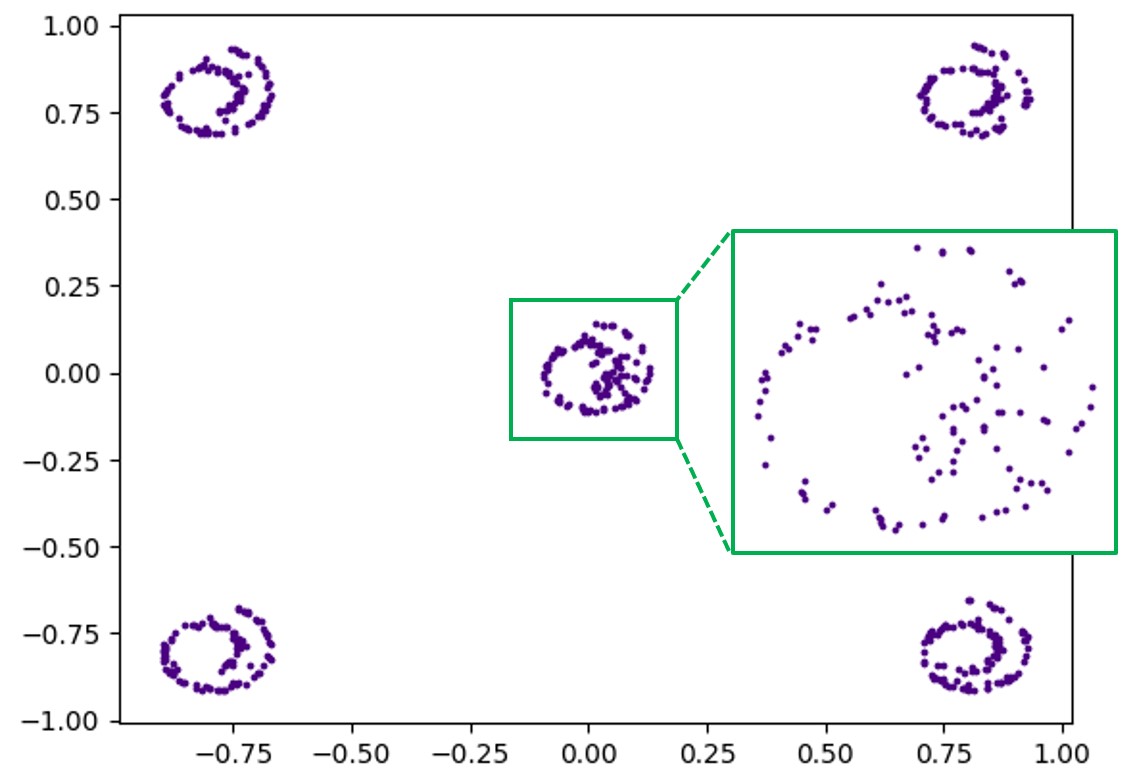}
    \label{fig:cld_2d_swiss_sample}
    \end{minipage}%
    }%
    \subfigure[HOLD]{
      \begin{minipage}[t]{0.5\linewidth}
      \centering
      \includegraphics[width=1.6in]{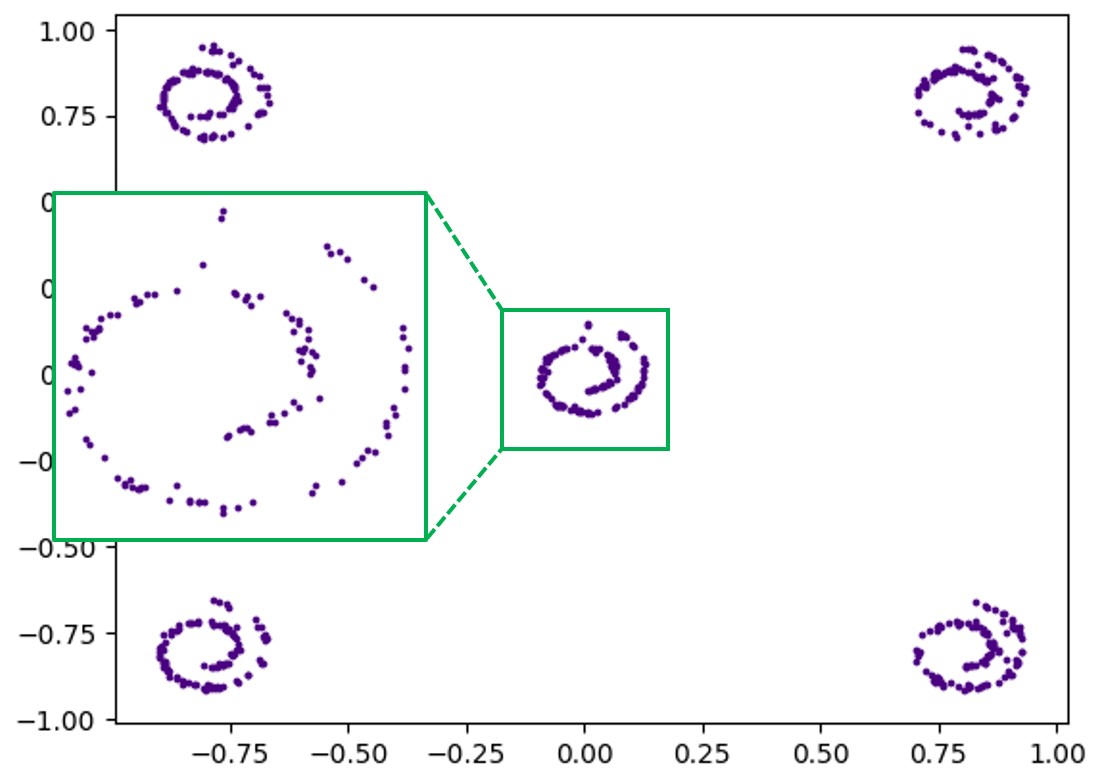}
      \label{fig:hold_2d_swiss_sample}
      \end{minipage}%
      }%
  \centering
  \caption{Comparison of 2D multi-Swiss rolls generated 
  by three DGMs based on VP SDE, CLD and HOLD.}
 \label{fig:2d_swiss_sample}
  \end{figure}

Looking at Eq.~(\ref{eq:third_order_langevin}) from an intuitive 
point of view, 
random noise can directly affect $\rvs_t$, while $\rvs_t$ can only
affect 
$\rvq_t$ indirectly through $\rvp_t$, so it can be expected that 
$\rvq_t$ is 
relatively smooth.
The trajectory $\rvq_t$ under these third-order dynamics have 
two additional orders of smoothness compared to the Wiener 
process $\rvw_t$. They are smoother than the corresponding 
trajectories under underdamped or critically damped Langevin 
dynamics~\cite{mou2021high,dockhorn2021score}. This higher 
smoothness degree can better control 
the discretization error.
Figure~\ref{fig:hold_1d_diffusion_qps} shows a one-dimensional 
example of HOLD. It can be seen that $\rvq$ can converge faster 
than $\rvp$ and $\rvs$.
It has been proved by~\cite{mou2021high} that certain discretization schemes
of Eq.~(\ref{eq:third_order_langevin}) can generate data from the 
%canonical distribution in $O (\frac{d^{1/4}}{\epsilon^{1/2}})$ steps, 
canonical distribution in $O (d^{1/4}/\epsilon^{1/2})$ steps,
which is (smoother and) exponentially faster than all the sampling 
scheme based on Eq.~(\ref{eq:diffusion_sde}), e.g. variance exploding (VE)
or variance preserving (VP) SDEs~\cite{song2020score}. 
Figures~\ref{fig:2d_swiss_paths} and~\ref{fig:2d_swiss_sample} 
show the generation of a complex 
multiple 2-dimensional Swiss rolls. Under the 
same conditions, it can be seen that HOLD has 
a better synthesis effect and a more stable generation 
speed than CLD. For example, Figure~\ref{fig:2d_swiss_paths} 
shows that at step 
699, it has been able to see that the sampling points 
have been gathered around the five cluster centers in HOLD, while 
the sampling points in CLD still look completely noisy.

\subsection{Block Coordinate Score Matching}

The key to using SDE for generation is the estimation of the 
score function, and the current best estimation method is based 
on DSM loss~(\ref{eq:dsm_loss}).
In the derivation of DSM loss for HOLD-based DGM, 
% $G(t) G(t)^\top \otimes \mI_d$ 
% has zero as almost all of its elements except 
% the sub-matrix $2\xi L^{-1} \otimes \mI_d$ in the lower right 
% corner. 
most of the elements in the matrix $G(t) G(t)^\top \otimes \mI_d$ 
are 0, except for 
the sub-matrix $2\xi L^{-1} \otimes \mI_d$ in the lower right corner.
So in $\nabla_{\rvx_t}$ only the derivatives with respect to
the entries in
$\rvs_t$ are not zero. 
% Thus the DSM loss for HOLD becomes the 
% following
% %
% \begin{align} 
% &\gL_{\mathrm{DSM}} \coloneqq \E_{\rvx_0 \sim p(\rvx_0), \rvx_t \sim p(\rvx_t \mid \rvx_0, t)} \nonumber\\&
% \left[\norm{\nabla_{\rvs_t} \log p(\rvx_t \mid \rvx_0, t) - \mathfrak{S}_{\theta}(\rvx_t, t)}_2^2\right].
% \end{align}\label{eq:hold_dsm_loss}
%
% If we assume $\vmu_0=(\rvq_0, \bm{0}_d, \bm{0}_d)$, and 
% $\Sigma_0^{qq}=0, \Sigma_0^{pp}=\alpha L^{-1} , \Sigma_0^{ss}=\alpha L^{-1} $ (
% $\mSigma_0 = \mathrm{diag}(\Sigma_0^{qq}, \Sigma_0^{pp}, \Sigma_0^{ss}) \otimes \mI_d$
% ), where $\alpha \ll  1$. Then $p(\rvx_t \mid \rvx_0, t) = p(\rvx_t \mid \rvq_0, t)$.
% Then we have

Since the numerical value of the score function 
$\nabla_{\rvs_t} \log p(\rvx_t \mid \rvx_0, t)$ in the DSM loss 
is not within 
a  bounded domain, and its estimation is often unstable, 
especially as the time $t$ tends to zero. 
By Eq.~(\ref{eq:transition_density}) we have  
$\nabla_{\rvx_t} \log p(\rvx_t \mid \rvx_0, t)=- \mL_t^{-\top} \rvepsilon_{3d}$, 
where $\rvepsilon_{3d} \sim \gN(\bm{0}, \mI_{3d})$ and 
$\mSigma_t = \mL_t \mL_t^\top$ is the Cholesky factorization 
of the covariance matrix $\mSigma_t$. The noise $\rvepsilon_{3d}$
is confined to a small bounded cubic almost all the time.
Therefore it is better to predict the noise than the score itself,
which has also been proved in the experiments 
by~\cite{kim2021score} and~\cite{song2020improved}.
Methods based on noise prediction always lead to 
better Frechet inception distance 
(FID)~\cite{ho2020denoising,rombach2022high}.
In experiments and applications, 
it is assumed $\vmu_0=(\rvq_0, \bm{0}_d, \bm{0}_d)$, and 
$\Sigma_0^{qq}=0, \Sigma_0^{pp}=\alpha L^{-1} , \Sigma_0^{ss}=\alpha L^{-1} $ (
$\mSigma_0 = \mathrm{diag}(\Sigma_0^{qq}, \Sigma_0^{pp}, \Sigma_0^{ss}) \otimes \mI_d$
), where $\alpha \ll  1$. Then $p(\rvx_t \mid \rvx_0, t) = p(\rvx_t \mid \rvq_0, t)$.
By reparameterization of $\nabla_{\rvs_t} \log p_t(\rvx_t \mid \cdot)= -\ell_t \rvepsilon_{2d:3d}$,  the 
DSM loss~(\ref{eq:dsm_loss}) is reformulated as
\begin{equation} 
  \E_{\rvx_0 \sim p(\rvx_0), \rvepsilon_{3d} \sim \gN(\bm{0}, \mI_{3d})}\left[ 
  \ell_t^2  
   \norm{ \rvepsilon_{2d:3d} + \mathfrak{E}_{\theta}(\vmu_t + \mL_t \rvepsilon_{3d}, t)}_2^2\right]. \nonumber
\end{equation}\label{eq:repara_hold_dsm_loss}
where $\mathfrak{E}_{\theta}(\vmu_t + \mL_t \rvepsilon_{3d}, t)=\mathfrak{S}_{\theta}(\vmu_t + \mL_t \rvepsilon_{3d}, t)/\ell_t$, and
$1/\ell_t$ is
\begin{align}
\sqrt{\Sigma^{ss}_t - \frac{(\Sigma^{sq}_t)^2}{\Sigma^{qq}_t} - \left(\Sigma^{pp}_t - \frac{\left(\Sigma^{pq}_t\right)^2}{\Sigma^{qq}_t}\right)^{-1}
  \left(\Sigma^{sp}_t - \frac{\Sigma^{sq}_t\Sigma^{pq}_t}{\Sigma^{qq}_t} \right)^2}. \nonumber
\end{align}
It is obvious that  $\vmu_t$, $\mL_t$, and $\Sigma_t$s all have $\rvx_0$ in their argument.

For general high-order dynamics, there are 
multiple variables (several blocks of coordinates), e.g. 
for third-order Langevin dynamic 
system in Eq.~(\ref{eq:third_order_langevin}), only one 
block $\rvq$ to represent the variables of interest and other two 
blocks $\rvp$ and $\rvs$ are introduced just to allow the two 
Hamiltonian dynamics to operate. There is no obvious reason that 
we need to denoise all blocks~\cite{dockhorn2021score}. 
Thus in this paper we propose
block coordinate score matching (BCSM)
to 
denoise only initial distribution $\rvb_0$ for  part  of 
all blocks $\rvx_0$, and 
marginalize over the distribution $p(\rvx_0 \backslash \rvb_0)$ of
remaining variables, which result in
 \begin{align} 
  \gL_{\mathrm{BCSM}} \coloneqq \E_{t \sim \gU[0, T]} 
  \lambda(t)   
  \E_{\rvb_0 \sim p(\rvb_0), \rvx_t \sim p(\rvx_t \mid \rvb_0, t)}
  \left[\norm{\nabla_{\rvs_t} \log p(\rvx_t \mid \rvb_0, t) - \mathfrak{S}_{\theta}(\rvx_t, t)}_2^2 \right].
  % &\gL_{\mathrm{BCSM}} \coloneqq \E_{t \sim \gU[0, T]} 
  % \lambda(t)   
  % \E_{\rvb_0 \sim p(\rvb_0), \rvx_t \sim p(\rvx_t \mid \rvb_0, t)} \nonumber\\&
  % \left[\norm{\nabla_{\rvs_t} \log p(\rvx_t \mid \rvb_0, t) - \mathfrak{S}_{\theta}(\rvx_t, t)}_2^2 \right].
  \label{eq:bcsm_loss}
\end{align}
BCSM is a flexible objective, and the only requirement for 
$\rvb_0$ is that it must contain block
$\rvq_0$, 
which is the variable we are interested in.
BCSM is also equivalent to SM, which is proved in the 
App.~\ref{sec:app_bcsm}.
After re-parameterization, the BCSM is 
\begin{align} 
  \E_{t \sim \gU[0, T]} 
 \lambda(t)   
  \E_{\rvb_0 \sim p(\rvb_0), \rvx_t \sim p(\rvx_t \mid \rvb_0, t)} 
  \left[ \norm{-\ell^{BCSM}_t \rvepsilon_{2d:3d} - \mathfrak{S}_\vtheta(\rvx_t, t)}_2^2 \right],
%   & \E_{t \sim \gU[0, T]} 
%  \lambda(t)   
%   \E_{\rvb_0 \sim p(\rvb_0), \rvx_t \sim p(\rvx_t \mid \rvb_0, t)}  \nonumber\\&
%   \left[ \norm{-\ell^{BCSM}_t \rvepsilon_{2d:3d} - \mathfrak{S}_\vtheta(\rvx_t, t)}_2^2 \right],
  \label{eq:bcsm_loss_reparam}
\end{align}
which is the format used in our implementation.
The intuition behind BCSM is that the distribution of  
$q_0$ is complex and unknown, 
and the distribution of $s_0$ is known, easy and fixed. 
That is to say, BCSM is easier to optimize, and 
ablation experiments also show that BCSM is better than the original DSM.

\section{Lie-Trotter Sampler (for HOLD-DGM)}

Methods of sampling  from a continuous diffusion generation 
model defined by an SDE are generally divided into two categories, 
one is based on the corresponding time-reverse SDE, e.g. 
Euler-Maruyama (EM)~\cite{song2020score}, GGF~\cite{jolicoeur2021gotta}, 
the other 
is based on the ODE flow corresponding to the reverse SDE, 
e.g. RK45~\cite{dormand1980family}, DEIS~\cite{zhang2022fast}, 
DPM-Solver~\cite{lu2022dpm}, GENIE~\cite{dockhorn2022genie}, etc. 
Among them, 
there are many acceleration algorithms based on ODE flow, because 
there is no random item, the direction 
of the path can be more accurately predicted.

This paper proposes a new sampling method inspired by 
molecular dynamics and 
thermodynamics~\cite{tuckerman2010statistical,leimkuhler2015molecular}.
The sampling problem is solved by the 
Lie-Trotter method, also known as the split 
operator method~\cite{trotter1959product}.
The principle of the Lie-Trotter method is to decompose 
a complex operator into two or more operators that
 are easy to calculate and iterate repeatedly. 
 Below we introduce how this principle is applied 
 to the sampling of DGM based on HOLD.%~\cite{leimkuhler2015molecular}

After we obtain the optimal predictor 
$\mathfrak{S}_\vtheta(\rvx_t, t)$ 
for score $\nabla_{\rvs_t} \log p(\rvx_t,t)$, 
which can be plugged into the backward HOLD for denoising the prior 
distribution into the target data.
The backward or the generative HOLD is approximated with 
\begin{align}
 d\rvq_t & = -\rvp_t dt, \nonumber\\
 d\rvp_t & =  \rvq_t  dt  -  \gamma\rvs_t dt, 
 \label{eq:generative_HOLD}\\
 d\rvs_t & =  \gamma\rvp_tdt + \xi\rvs_t dt 
 + 2\xi L^{-1} \mathfrak{S}_\vtheta(\rvx_t, t)dt
 + \sqrt{2\xi L^{-1}} d\bar\rvw_t. \nonumber
\end{align}

We regard the right side of the backward 
HOLD~(\ref{eq:generative_HOLD}) as the 
superposition of 
multiple 
different vector fields, and the principle of decomposition is 
that the SDE 
or differential equation (DE) corresponding to individual vector field has an exact 
solution.
The exact solution here means that it is possible to sample 
from the 
distribution corresponding to this component.
Assume $p(\rvx_0,0)$ is an initial distribution in the phase space,
then the propagation is determined 
$p(\rvx_t,t)=e^{t\mathcal{L}_{H}^\dagger}p(\rvx_0,0)$ 
where
$\mathcal{L}_{H}^\dagger$ is the Fokker-Planck operator 
for time reverse HOLD~(\ref{eq:generative_HOLD}) 
and $p(\rvx_t,t)$  is 
the evolved distribution at time $t$.
From experience in solving higher-order SDE 
equations~(\ref{eq:nth_order_langevin}), 
in particular the previous 
estimates of transition probability kernels for third-order 
Langevin dynamical 
systems~(\ref{eq:third_order_langevin}), we have exact closed-form 
solutions for the 
superimposition of
Ornstein-Uhlenbeck process 
and the sum of the two Hamiltonian components. 
So we first isolate the following SDE from the right 
side of the backward Eq.~(\ref{eq:generative_HOLD})
% % 
% \begin{equation} 
%   \begin{pmatrix} d\rvq_t \\ d\rvp_t \\ d\rvs_t \end{pmatrix} 
%   =\begin{pmatrix}  \bm{0}_d \\ \rvq_t \\\bm{0}_d \end{pmatrix} dt 
%   +  \begin{pmatrix}  -\rvp_t \\ \bm{0}_d \\ \gamma \rvp_t \end{pmatrix} dt +
%   \begin{pmatrix}\bm{0}_d \\ -\gamma \rvs_t \\  -\xi\rvs_t\end{pmatrix} dt +
%   \begin{pmatrix}\bm{0}_d \\ \bm{0}_d \\ \sqrt{2\xi L^{-1}} d\bar \rvw_t \end{pmatrix}.
% \end{equation}
%
% 
\begin{equation} 
  \begin{pmatrix} d\rvq_t \\ d\rvp_t \\ d\rvs_t \end{pmatrix} 
=\begin{pmatrix}  -\rvp_t \\ \rvq_t-\gamma \rvs_t \\\gamma \rvp_t -\xi\rvs_t \end{pmatrix} dt 
  +  
\begin{pmatrix}\bm{0}_d \\ \bm{0}_d \\ \sqrt{2\xi L^{-1}} d\bar \rvw_t \end{pmatrix}.
\label{eq:hold_vector_a}
\end{equation}
It is different from Eq.~(\ref{eq:third_order_langevin}), 
but very similar, 
so many calculations related to Eq.~(\ref{eq:third_order_langevin}) 
can be extended to Eq.~(\ref{eq:hold_vector_a}).
And the remaining vector field is
\begin{equation} 
  \begin{pmatrix} d\rvq_t \\ d\rvp_t \\ d\rvs_t \end{pmatrix} 
  =
  \begin{pmatrix}\bm{0}_d \\ \bm{0}_d \\ 
    2\xi \left[L^{-1}\mathfrak{S}_\vtheta(\rvx_t, t) +\rvs_t \right] \end{pmatrix}dt.
\label{eq:hold_vector_b}
\end{equation}
Let $\mathcal{L}_{A}^\dagger$ and $\mathcal{L}_{B}^\dagger$ 
be the Fokker-Planck operators
of SDE~(\ref{eq:hold_vector_a}) and 
DE~(\ref{eq:hold_vector_b}) respectively.
Thus we have 
$\mathcal{L}_{H}^\dagger=\mathcal{L}_{A}^\dagger+\mathcal{L}_{B}^\dagger$ 
and 
$e^{t\mathcal{L}_{H}^\dagger}=e^{t\left(\mathcal{L}_{A}^\dagger+\mathcal{L}_{B}^\dagger\right)}$.
It is easy to verify that $\mathcal{L}_{A}^\dagger$ 
and $\mathcal{L}_{B}^\dagger$ 
cannot be exchanged (commute). 
Fix a constant valued function $p$ and we can see that 
$\mathcal{L}_{A}^\dagger\mathcal{L}_{B}^\dagger(p)\neq 0$, 
while $\mathcal{L}_{B}^\dagger\mathcal{L}_{A}^\dagger(p)=0$.
% Thus $\mathcal{L}_{A}^\dagger$ and $\mathcal{L}_{B}^\dagger$ are 
% noncommuting operators and the operator
% difference 
% $\mathcal{L}_{A}^\dagger\mathcal{L}_{B}^\dagger-\mathcal{L}_{B}^\dagger\mathcal{L}_{A}^\dagger$ 
% is an object that arises frequently both in classical and
% quantum mechanics and is known as the commutator between the operators.
% Since $\mathcal{L}_{A}^\dagger$ and $\mathcal{L}_{B}^\dagger$ 
% are not commutative, 
Thus the classical propagator 
$e^{t\left(\mathcal{L}_{A}^\dagger+\mathcal{L}_{B}^\dagger\right)}$ 
cannot be separated into a simple product
$e^{t\mathcal{L}_{A}^\dagger}e^{t\mathcal{L}_{B}^\dagger}$.
This is unfortunate because in many cases the effect of the 
individual operators 
$e^{t\mathcal{L}_{A}^\dagger}$ and $e^{t\mathcal{L}_{B}^\dagger}$ 
on the phase space vector can be precisely 
evaluated. 
So it would be useful if the commutator could be represented 
by these 
two factors.
In fact, there is a way to do this using an important 
theorem called Trotter's theorem~\cite{trotter1959product}. 
The theorem states 
that for two operators $\mathcal{L}_{A}^\dagger$ 
and $\mathcal{L}_{B}^\dagger$
\begin{equation} 
e^{t\mathcal{L}_{A}^\dagger+t\mathcal{L}_{B}^\dagger}
=\lim_{P \to \infty} 
\left[e^{\frac{t\mathcal{L}_{A}^\dagger}{2P}}
e^{\frac{t\mathcal{L}_{B}^\dagger}{P}}
e^{\frac{t\mathcal{L}_{A}^\dagger}{2P}}\right]^P
\label{eq:strang_split_form}
\end{equation}
where $P$ is a positive integer. 
Thus in a word, we apply symmetric Trotter theorem to 
classical propagator 
generation. 
Eq.~(\ref{eq:strang_split_form}) is commonly 
referred to as the symmetric
Trotter theorem or Strang splitting 
formula~\cite{strang1968construction}, which 
states that we can propagate a classical 
system using the separate
factor $e^{\frac{t\mathcal{L}_{A}^\dagger}{2P}}$ and 
$e^{\frac{t\mathcal{L}_{B}^\dagger}{P}}$ exactly for 
a finite time 
$t$ in the limit that $P \to \infty$.
The propagation is determined by each of the two operators 
applied successively to $\rvx$ and $\rvs$.
The above analysis demonstrates how we can obtain 
the acceleration Verlet algorithm
via the powerful formalism provided by the Trotter 
factorization scheme. 

  \vspace{-0.2cm}

% Let the number of steps we take to go to infinity and the time 
% step go to zero. Of course,
% this is not practical, but if we do not take these limits, 
% then Eq.~(\ref{eq:strang_split_form}) leads to a
% useful approximation for classical propagation.
% Note that for finite $P$, Eq.~(\ref{eq:strang_split_form}) 
% implies an approximation. 

In the calculation process 
of $e^{\frac{t\mathcal{L}_{A}^\dagger}{2P}}$, 
the mean and variance 
after $\frac{t}{2P}$ time need to be calculated.
By solving the generative HOLD~(\ref{eq:generative_HOLD}), 
$e^{\mathcal{L}_{A}^\dagger\frac{t}{2P}} \rvx_t$
is a Gaussian distribution with mean $\vmu_t$ and 
variance $\mSigma_t$
by integrating
\begin{align} 
  \frac{d\vmu_t}{dt} &= \vl(t) \vmu_t, \label{app:ode_mean_sampling} \\
  \frac{d\mSigma_t}{dt} &=\vl(t) \mSigma_t + \left[ \vl(t) \mSigma_t \right]^\top + 
  \mM(t) \mM(t)^\top . \label{app:ode_var_sampling}
 \end{align}
 over $[t, t+\frac{t}{2P}]$, where $\vl(t)$ and $\mM(t)$ are
\begin{align}
  \begin{pmatrix} 0 &-1&0\\ 
    1 &0&-\sqrt{10}\\ 
    0 &  \sqrt{10} & -6 \end{pmatrix} \otimes \mI_d, \quad
\begin{pmatrix} 0 &0 &0\\ 
    0 &0 &0\\ 
    0  & 0 &\sqrt{12 L^{-1}} \end{pmatrix} \otimes \mI_d,\nonumber
\end{align}
respectively.
Please refer to the App.~\ref{sec:app_calc_op_a} for the 
details of calculation 
process and results.

% \begin{wrapfigure}{r}{0.4\textwidth}
%   \vspace{-20pt}
%   \begin{center}
%   \includegraphics[width=0.4\textwidth]{celeb_sample.png}
%   \end{center}
%   \vspace{-5pt}
%   \caption{
%     Generated CelebA-HQ-256 samples.
%   }
%   \label{fig:celeb_sample}
%   \vspace{-10pt}
% \end{wrapfigure}

For the second operator $e^{\mathcal{L}_{B}^\dagger\frac{t}{P}}$, which involves a derivative with 
respect to acceleration variable acts on the
$\rvx_t$ appearing in both components of the vector 
appearing on the right side of Eq.~(\ref{eq:hold_vector_b}).
We use one step Euler method to do the updating (solving)
of Eq.~(\ref{eq:hold_vector_b}).
% % 
% \begin{equation} 
%   e^{\mathcal{L}_{B}^\dagger\frac{t}{P}} \begin{pmatrix} \rvq_t \\ \rvp_t \\ \rvs_t \end{pmatrix} 
%   = \begin{pmatrix} \rvq_t \\ \rvp_t \\ \rvs_t \end{pmatrix}  +
%   \begin{pmatrix}\bm{0}_d \\ \bm{0}_d \\ 2\xi \left[L^{-1}\mathfrak{S}_\vtheta(\rvx_t, t) +\rvs_t \right] \end{pmatrix}\frac{t}{P}.
% \nonumber
% \end{equation}
% %
Whether it is step~(\ref{eq:hold_vector_a}) or 
step~(\ref{eq:hold_vector_b}), each iteration of the 
Lie-Trotter algorithm 
is accurate sampling, and the mixing speed of 
HOLD is also very fast, so the sampling algorithm can 
converge quickly, thus sampling at a fast speed.
Experimental results in Section~\ref{sec:exp_img_gen}
also show the effectiveness of the Lie-Trotter method.
% The probability flow ODE of HOLD is
% %
% \begin{equation} 
%   \begin{cases}
%     d\rvq_t & = \rvp_t dt, \\
%     d\rvp_t & =  -\rvq_t  dt  +  \gamma\rvs_t dt, \\
%     d\rvs_t & =  -\gamma\rvp_tdt - \xi\rvs_t dt  + \sqrt{2\xi L^{-1}} d\rvw_t
%   \end{cases}\label{eq:third_order_ode_prob_flow}
%   \end{equation}
%   %

\section{Image Restoration with HOLD}

\section{Experiments}

\begin{table}%{r}{6.3cm}
  \centering
  \caption{\small Unconditional CIFAR-10 generative performance.} 
  \label{tab:cifar10_main}
  \scalebox{0.65}{\begin{tabular}{c l c c}
  \toprule
 Model & NLL$\downarrow$ & FID$\downarrow$ \\
  \midrule
   HOLD-DGM (Prob. Flow) \textit{(ours)} & $\leq$2.94 & 1.88 \\
   HOLD-DGM (SDE) \textit{(ours)} & - & 1.85 \\
  \midrule
  PFGM++~\cite{xu2023pfgm++}  & - &1.91 \\
  EDM~\cite{karraselucidating}  & - & 1.97 \\
  CLD-SGM (Prob. Flow)  & $\leq$3.31 & 2.25 \\
   CLD-SGM (SDE)  & - & 2.23 \\
   DDPM++, VPSDE (SDE)~\cite{song2020score} & - & 2.41 \\
   DDPM \cite{ho2020denoising} & $\leq$3.75 & 3.17 \\
   Likelihood SDE \cite{song2021maximum} &  2.84 & 2.87 \\
   DDIM (100 steps)  \cite{song2020denoising} & - & 4.16\\
   FastDDPM (100 steps) \cite{kong2021fast} & - &  2.86\\
   Improved DDPM \cite{nichol2021improved} & 3.37 &2.90\\
   UDM \cite{kim2021score} &  3.04 &2.33\\
  \midrule
 StyleGAN2 w/ ADA \cite{karras2020training} & - & 2.92\\
   Glow \cite{kingma2018glow} & 3.35 & 48.9\\
   Residual Flow \cite{chen2019residual} & 3.28 & 46.37\\
 DC-VAE~\cite{parmar2021dual} & - & 17.90 \\
 VAEBM \cite{xiao2020vaebm} & - & 12.2\\
   Recovery EBM \cite{gao2020learning} & 3.18 & 9.58\\
  \bottomrule
  \end{tabular}}
  \end{table}
  \vspace{-0.2cm}

HOLD-DGM is a general data generation model that 
can synthesize data of arbitrary modality. In order 
to verify its effectiveness, in this paper we use 
HOLD-DGM to generate a variety of image data.
Similar to many generative models based on score and SDE, 
only one neural network $\mathfrak{S}_\vtheta(\rvx_t, t)$ 
is needed in the whole system to 
predict the score. 
In this paper, we consistently use 
NCSN++~\cite{song2020improved,song2021maximum} 
with 9 input channels 
(for data, velocity, and acceleration) instead of 3 
for  $\mathfrak{S}_\vtheta(\rvx_t, t)$.

From Section~\ref{sec:told}, it can be seen that HOLD-DGM does 
not have many hyperparameters, and in order to make the 
base of the mean and variance of the transition probability 
simple and elegant, several hyperparameters are fixed in 
all experiments, such as 
$\gamma{=}\sqrt{10}$ and $\xi{=}6$.
%which are fixed in all experiments. 
In this way, the adjustable hyperparameters in the entire 
HOLD-DGM are only $\alpha$ (the variance scaling of the initial 
velocity and acceleration   
distribution) 
and $L$, except for the parameters in 
the neural network.
All HOLD-DGM are trained under the proposed BCSM objective, 
and we found that as long as $\rvb_0$ is not equal to 
\{$\rvq_0$, $\rvp_0$, $\rvs_0$\}, the basic performance is similar. 
We use $\rvb_0=\{\rvq_0\}$ in all experiments.
The BCSM 
objective is used with the weighting 
$\lambda(t){=}(\ell^{BCSM}_t)^{-2}$, 
which promotes image quality.

For the sampling algorithm, to highlight our proposed 
(Lie-Trotter) LT algorithm, we tried two other types 
of sampling algorithms, 
including \textbf{(i)} Probability flow 
using a Runge--Kutta method; reverse-time generative 
HOLD sampling using 
\textbf{(ii)} EM.
For EM and LT, we use evaluation times 
chosen according to 
a quadratic function, like previous 
work~\cite{kong2021fast,watson2021learning}.

In order to evaluate the quality of the images generated by 
the model trained on CIFAR-10, two metrics are used in this paper, 
one is FID with 50,000 
samples~\cite{heusel2017gans}
and the other is negative 
log-likelihood~(NLL). Similar to most SDE-based generative 
models, HOLD-DGM has no way to accurately calculate NLL, 
so this paper uses an upper bound to approximate.
This upper bound is obtained by
${-}\log p(\rvq_0){\leq}{-} \E_{\rvp_0, \rvs_0\sim p(\rvp_0, \rvs_0)}\log p_\varepsilon(\rvq_0, \rvp_0, \rvs_0){-}H$ (refer
App.~\ref{sec:app_likelihood_comput} for derivation details), 
where $H$ is the entropy of $p(\rvp_0, \rvs_0)$ and $\log p_\varepsilon(\rvq_0, \rvp_0, \rvs_0)$ is 
an unbiased estimate of $\log p(\rvq_0, \rvp_0, \rvs_0)$ from the 
probability 
flow ODE~\cite{grathwohl2018ffjord,song2020score} of HOLD.
% %
% \begin{equation} 
%   \begin{cases}
%     d\rvq_t & = \rvp_t dt, \\
%     d\rvp_t & =  -\rvq_t  dt  +  \gamma\rvs_t dt, \\
%     d\rvs_t & =  -\gamma\rvp_tdt - \xi\rvs_t dt  + \sqrt{2\xi L^{-1}} d\rvw_t.
%   \end{cases}\label{eq:third_order_ode_prob_flow}
%   \end{equation}
%   %
% As in~\cite{vahdat2021score}, the 
% stochasticity of $\log p_\varepsilon(\rvq, \rvp, \rvs)$ prevents 
% us from 
% performing importance-weighted NLL estimation over the
%   velocity and acceleration distribution~\cite{burda2015importance}.
And since HOLD-DGM is not trained with the goal of minimizing NLL, 
its performance on NLL is not the best. This is a future work 
where we will try to train HOLD-DGM using maximum 
likelihood as the objective.

% \begin{wrapfigure}{r}{0.45\textwidth}
%   \vspace{-20pt}
%   \begin{center}
%   \includegraphics[width=0.4\textwidth]{cifar10_sample.png}
%   \end{center}
%   \vspace{-5pt}
%   \caption{
%     Generated CIFAR-10 samples without cherry-picking.
%   }
%   \label{fig:cifar10_sample}
%   \vspace{-10pt}
% \end{wrapfigure}

In order to compare the generation efficiency of 
different models, we also counted 
the number of function evaluations (NFEs, mainly score 
calculations), 
which is the most computationally resource-intensive part of the 
generation process.

We have done 1D data, 2D data, and high-dimensional data (image) 
generation experiments, and all experimental setting details 
are in App.~\ref{appsec:exp_details}. 
To save space, here we only 
introduce the effect for image generation.
\subsection{2D multi-swiss roll generation}

\subsection{Image Generation}
\label{sec:exp_img_gen}

Similar to~\cite{song2020improved} and~\cite{dockhorn2021score}, 
we also conduct 
experiments on the widely 
used CIFAR-10 unconditional image generation 
benchmark for verification.

\begin{figure}[th]%[htp]
  %\vspace{-0.4in}
  \centering
  %\begin{center}
  \hspace{-5mm}
  \includegraphics[width=1.0\linewidth]{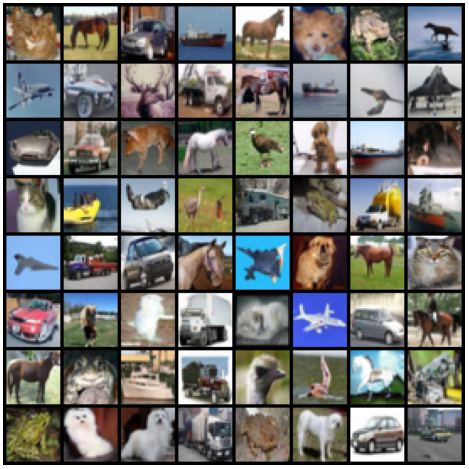}
  \hspace{-5mm}
  %\end{center}
  %\vspace{-2.5mm}
  \caption{
    Generated CIFAR-10 samples without cherry-picking.
  }
  \label{fig:cifar10_sample}
  %\vspace{-2.5mm}
  \end{figure}

First of all, if the computing resource limitation during 
sampling is not considered, that is, there is no upper 
limit for NFE, we can achieve state-of-the-art performance. 
As shown in Table~\ref{tab:cifar10_main}, when 
using HOLD ODE 
stream sampling, FID is 2.22, and when using reverse HOLD 
sampling, FID is 2.20. Although our NLL is not optimal, 
it is also in a very competitive position from the perspective 
of the upper 
bound~\cite{dockhorn2021score,vahdat2021score}.
This shows that our HOLD-DGM generates data with no 
fewer modes than other best generative models.
In the case of considering computation cost, i.e. 
under the same NFE budget
and capacity of the model, our HOLD-DGM has a better 
performance than other models, as shown in Table~\ref{tab:nfe_results}.
Due to the special form of HOLD, our proposed LT 
sampling algorithm has customized excellent performance. 
It can be seen that LT is much better than EM, 
and LT is also much better than SSCS 
algorithm proposed by~\cite{dockhorn2021score}.
Figure~\ref{fig:cifar10_sample} shows some generated CIFAR-10 samples.

\begin{figure}[H]
    %\vspace{-0.4in}
    \centering
    %\begin{center}
    \hspace{-5mm}
    \includegraphics[width=0.8\linewidth]{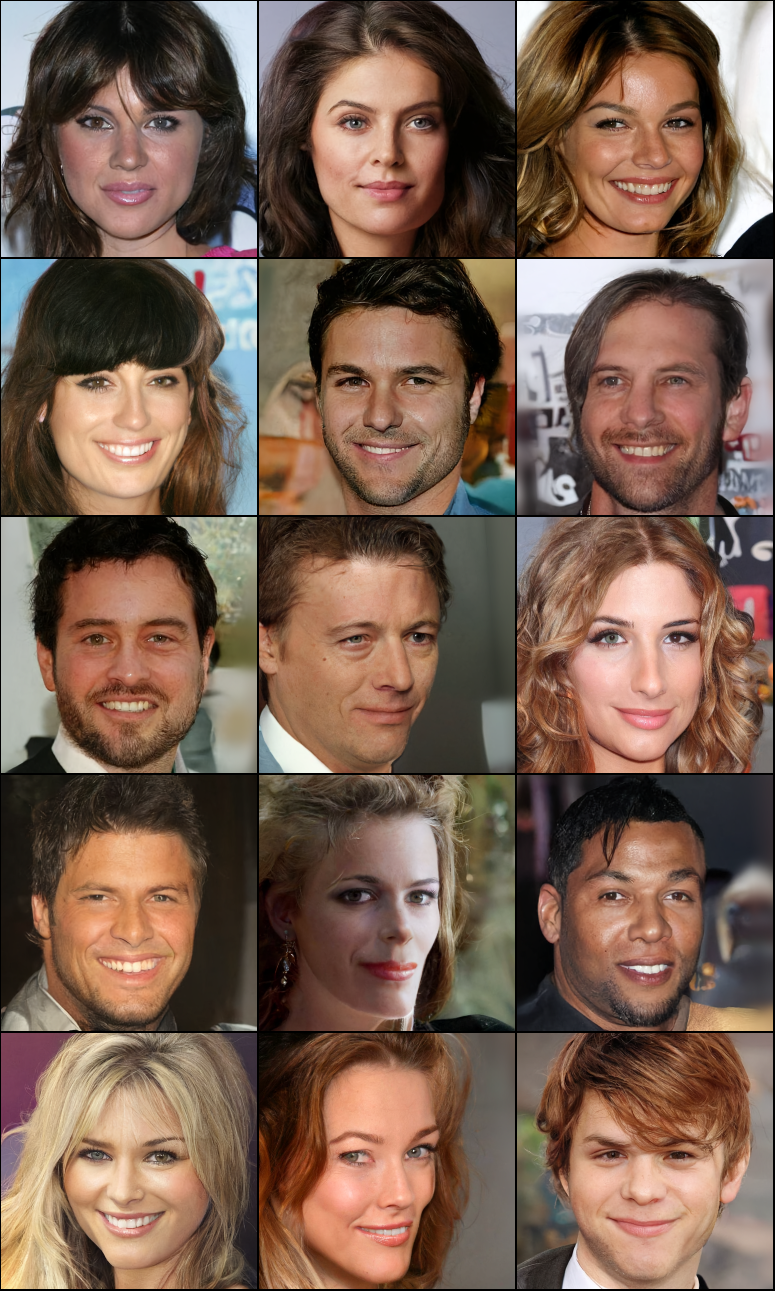}
    \hspace{-5mm}
    %\end{center}
    %\vspace{-2.5mm}
    \caption{
      Generated CelebA-HQ-256 samples.
    }
    \label{fig:celeb_sample}
    %\vspace{-2.5mm}l
    \end{figure}

HOLD-DGM not only has very good performance on low-resolution 
CIFAR-10, 
but it can also synthesize high-resolution images. 
Figure~\ref{fig:celeb_sample} shows 
the images generated by HOLD-DGM after training on 
CelebA-HQ~\cite{karras2018progressive} 
with a resolution of $256\times 256$. It can be seen that our model 
is very good in terms of the authenticity and diversity (that is,
 coverage) of the faces. There are also more sample images 
 in App.~\ref{appsec:gen_image}.

  \vspace{-0.5cm}

\begin{table}%{r}{7.7cm}
    \centering
    \caption{Performance comparison under the 
    same NFEs (Results other than HOLD come 
    from~\cite{dockhorn2021score}).} 
\label{tab:nfe_results}
\scalebox{0.75}
{\begin{tabular}{l l c c c c c}
        \toprule
        & & \multicolumn{5}{c}{FID at $n$ function evaluations $\downarrow$} \\
        Model &Sampler & $n{=}50$ & $n{=}150$ &  $n{=}500$ & $n{=}1000$ & $n{=}2000$ \\
        \midrule 
        HOLD & LT & \textbf{15.1} & \textbf{2.67} &  \textbf{2.07} & \textbf{1.96} & \textbf{1.85}\\
        HOLD & EM & 38.2 & 5.60 &   2.24  & 2.18 & 2.12\\
        \midrule 
        CLD & SSCS & 20.5 & 3.07 &  2.25 & 2.30 & 2.29\\
        VPSDE & EM & 28.2 & 4.06 & 2.47 & 2.66 & 2.60 \\
        VESDE & PC & 460 & 216 &  3.75  & 2.43 & 2.23 \\
        \bottomrule
    \end{tabular}}
\end{table}
%\vspace{-0.2cm}

\subsection{Ablation Studies}

We conducted ablation studies to prove that HOLD is very 
effective in 
DGM, and it is not sensitive to hyperparameters.
The network configuration used here is the same as that used above.
We did this research on three aspects in HOLD-DGM, namely 
the hyperparameters $L$ and $\alpha$ 
sampling algorithm.
$L$ is mainly in the sub-equation of 
acceleration in HOLD~(\ref{eq:third_order_langevin}), which 
is used to control the influence of Wiener process on 
acceleration. The smaller $L$ is, the greater the influence 
of Wiener process on acceleration, so that the inflow acceleration 
and speed will be faster.
$\alpha$ is the initial velocity and acceleration variance scalings. 
Table~\ref{tab:hyp_ablation} shows the performance of HOLD-DGM 
under different $L$ and $\alpha$. 
Intuitively, the results are not much different. But when 
$L$ is relatively small, the performance of FID will be better.
The situation is very similar for $\alpha$, 
a small $\alpha$ will have slightly better performance. That is 
to say, at the beginning of data diffusion, the speed and 
acceleration should be set smaller, so that a better 
FID can be obtained

The experimental results also show that the performance 
  of HOLD-DGM is not sensitive to hyperparameters $L$ and $\alpha$ , 
  which is a good property and is easy to transplant to 
  other generation tasks, such as 3D data.

\begin{table}[th]
  %\begin{wraptable}{r}{6.3cm}
    \caption[hyp_ablation]{FID performance for 
    different $L$ and $\alpha$.}
    \label{tab:hyp_ablation}
    \centering
    \begin{tabular}{c|c|c|c}
    \hline
    \hline
  FID$\downarrow$ & $L=1.0$& $L=2.0$ & $L=4.0$\\
    \hline
    $\alpha=0.02$ &  1.89&  1.85&  1.95\\
    $\alpha=0.04$ & 1.92 &  1.90&  1.93\\
    $\alpha=0.08$ & 1.96 &  1.91&  2.00\\
    \hline
    \end{tabular}
 %   \end{wraptable}
    \end{table}

\section{Conclusion}
\label{sec:conclusion}

We introduce a new framework for score-based generative modeling 
with high-order Langevin dynamics. Our work enables a
deep understanding of SDE-based approaches, 
decoupling of data and Wiener processes,
smoother sampling path, 
flexible objective,
sampling with higher expressiveness, 
and brings new 
possibilities to the family of generative models based on SDE 
and score matching.

% \begin{ack}
% Use unnumbered first level headings for the acknowledgments. All acknowledgments
% More information about this disclosure can be found at: \url{https://neurips.cc/Conferences/2023/PaperInformation/FundingDisclosure}.

% Do {\bf not} include this section in the anonymized submission, only in the final paper. You can use the \texttt{ack} environment provided in the style file to autmoatically hide this section in the anonymized submission.
% \end{ack}

% \section{Supplementary Material}

% Authors may wish to optionally include extra information (complete proofs, additional experiments and plots) in the appendix. All such materials should be part of the supplemental material (submitted separately) and should NOT be included in the main submission.

%\section*{References}

\bibliographystyle{ieee_fullname}

% References follow the acknowledgments in the camera-ready paper. Use unnumbered first-level heading for
% the references. Any choice of citation style is acceptable as long as you are
% consistent. It is permissible to reduce the font size to \verb+small+ (9 point)
% when listing the references.
% Note that the Reference section does not count towards the page limit.
\medskip

{
\small

\bibliography{hold}
% [1] Alexander, J.A.\ \& Mozer, M.C.\ (1995) Template-based algorithms for
% connectionist rule extraction. In G.\ Tesauro, D.S.\ Touretzky and T.K.\ Leen
% (eds.), {\it Advances in Neural Information Processing Systems 7},
% pp.\ 609--616. Cambridge, MA: MIT Press.

% [2] Bower, J.M.\ \& Beeman, D.\ (1995) {\it The Book of GENESIS: Exploring
%   Realistic Neural Models with the GEneral NEural SImulation System.}  New York:
% TELOS/Springer--Verlag.

% [3] Hasselmo, M.E., Schnell, E.\ \& Barkai, E.\ (1995) Dynamics of learning and
% recall at excitatory recurrent synapses and cholinergic modulation in rat
% hippocampal region CA3. {\it Journal of Neuroscience} {\bf 15}(7):5249-5262.

}

%%%%%%%%%%%%%%%%%%%%%%%%%%%%%%%%%%%%%%%%%%%%%%%%%%%%%%%%%%%%

\section*{Supplementary Material}

\section{A Lemma}

We first introduce a lemma that will 
be used in some fact derivations in the following sections.

\begin{lem}
   \label{lem:ekp}
   (Exponential of Kronecker product) Let $\mA\in\R^{d \times d}$ and $\mI$ is identity matrix, then $\exp(\mA\otimes \mI)=\exp(\mA)\otimes \mI$.
   \end{lem}
   \begin{proof}
     First we have
     \begin{align*}
       \left(\mA\otimes \mI\right)\left(\mA\otimes \mI\right)  
       =  \begin{pmatrix}
         a_{11} \mI      & a_{12}\mI &  \dots & a_{1d}\mI \\
         a_{21}  \mI     & a_{22}\mI &  \dots & x_{2n} \mI\\
         \vdots & \vdots &\vdots & \vdots  \\
         a_{d1}  \mI     & a_{d2}\mI &  \dots & a_{dd}\mI
     \end{pmatrix}
     \begin{pmatrix}
       a_{11} \mI      & a_{12}\mI &  \dots & a_{1d}\mI \\
       a_{21}  \mI     & a_{22}\mI &  \dots & x_{2n} \mI\\
       \vdots & \vdots &\vdots & \vdots  \\
       a_{d1}  \mI     & a_{d2}\mI &  \dots & a_{dd}\mI
   \end{pmatrix}
       = \mA^2\otimes \mI.
     \end{align*}
     Then by induction on $n$, we obtain
     \begin{align*}
       \left(\mA\otimes \mI\right)^n = \mA^n\otimes \mI.
     \end{align*}
     Therefore
     \begin{align*}
       \exp\left(\mA\otimes \mI\right)=1+\sum_{n=1}^{\infty}\frac{\left(\mA\otimes \mI\right)^n}{n!}=1+\sum_{n=1}^{\infty}\frac{\mA^n\otimes \mI}{n!}
       =1+\sum_{n=1}^{\infty}\frac{\mA^n}{n!}\otimes \mI = \exp(\mA)\otimes \mI.
     \end{align*}
   \end{proof}

\section{High-Order Langevin Dynamics}
A pair of general linear forward diffusion and backward SDE is
\begin{align}
  d\rvx_t &= \vf(t)\rvx_t \, dt + \mG(t) \, d\rvw_t, \quad t \in [0, T]
\label{eq:app_sde}
\end{align}
and
\begin{align} 
    d\rvx_t &= \left[-\vf(t)\rvx_t + \mG(t) \mG(t)^\top 
    \nabla_{\rvx_t} \log p_{t}(\rvx_t)\right] \, dt + \mG(t) d\bar \rvw_t,
    \label{eq:app_rsde}
\end{align}
where 
$\vf(t), \mG(t) \in \R^{3d\times 3d}$,
 $\rvw_t$ is the standard 
 Wiener process, and $\bar \rvw_t$ is time-reverse Wiener process. 
Recall that the High-order (third-order) Langevin dynamics (HOLD) in 
the main paper is 
 \begin{equation} 
  \begin{split}
    d\rvq_t & = \rvp_t dt, \\
    d\rvp_t & =  -mL^{-1}\rvq_t  dt  +  \gamma\rvs_t dt, \\
    d\rvs_t & =  -\gamma\rvp_tdt - \xi\rvs_t dt  + \sqrt{2\xi L^{-1}} d\rvw_t.
   \label{eq:app_hold}
  \end{split}
\end{equation} 
We can write it in matrix form as Eq.~(\ref{eq:app_sde}), and it is
\begin{equation} 
  \begin{pmatrix} d\rvq_t \\ d\rvp_t \\ d\rvs_t \end{pmatrix} 
  =\begin{pmatrix} \rvp_t \\ -mL^{-1}\rvq_t \\\bm{0}_d \end{pmatrix} dt 
  +  \begin{pmatrix} \bm{0}_d \\ \gamma \rvs_t \\ -\gamma \rvp_t \end{pmatrix} dt +
  \begin{pmatrix}\bm{0}_d \\ \bm{0}_d \\ - \xi\rvs_t\end{pmatrix} dt +
  \begin{pmatrix}\bm{0}_d \\ \bm{0}_d \\ \sqrt{2\xi L^{-1}} d\rvw_t \end{pmatrix}.
  \label{eq:app_hold_matrix}
\end{equation}
Let $\rvx_t = (\rvq_t, \rvp_t, \rvs_t)^\top \in \R^{3d}$, then we 
have
\begin{align}
  \vf(t) \coloneqq  \begin{pmatrix} 0 &1&0\\ 
    -mL^{-1} &0&\gamma\\ 
    0 & - \gamma & -\xi \end{pmatrix} \otimes \mI_d,
    \qquad
  \mG(t) \coloneqq \begin{pmatrix} 0 &0 &0\\ 
    0 &0 &0\\ 
    0  & 0 &\sqrt{2\xi L^{-1}} \end{pmatrix} \otimes \mI_d,
    \label{eq:app_hold_matrix}
\end{align}
where $\otimes$ denotes the Kronecker product.
For the simplicity and elegance of the calculation, in this paper
we set $m=L$, $\xi=6$, and $\gamma=\sqrt{10}$
then the HOLD~(\ref{eq:app_hold}) is simplified 
into Eq.~(\ref{eq:app_sde}) with
% 
% \begin{align}
%   \vf(t) \coloneqq  \begin{pmatrix} 0 &1&0\\ 
%     -1 &0&\gamma\\ 
%     0 & - \gamma & -\xi \end{pmatrix} \otimes \mI_d,
%     \qquad
%   \mG(t) \coloneqq \begin{pmatrix} 0 &0 &0\\ 
%     0 &0 &0\\ 
%     0  & 0 &\sqrt{2\xi L^{-1}} \end{pmatrix} \otimes \mI_d,
% \end{align}
%
% In this paper, we let $\xi=6$ and $\gamma=\sqrt{10}$, so we have
% 
\begin{align}
   \vf(t) \coloneqq  \begin{pmatrix} 0 &1&0\\ 
     -1 &0&\sqrt{10}\\ 
     0 & - \sqrt{10} & -6 \end{pmatrix} \otimes \mI_d,
     \qquad
   \mG(t) \coloneqq \begin{pmatrix} 0 &0 &0\\ 
     0 &0 &0\\ 
     0  & 0 &\sqrt{12 L^{-1}} \end{pmatrix} \otimes \mI_d.
 \end{align}
 \subsection{Perturbation Kernel}

 For linear SDE, we can get the score by solving an associated
 deterministic ordinary differential 
 equation. The transition densities 
 $p(\rvx_t|\rvx_0)$ of the solution 
 process $\rvx_t$ for the SDE~(\ref{eq:app_hold}) 
 is the solution to the Fokker-Planck-Kolmogorov (FPK)
 equation~\cite{sarkka2019applied}
 \begin{align}
   \frac{\partial p(\rvx_t,t)}{\partial t}  = 
   -\sum_{i=1}^d\frac{\partial \left[ (\vf(t)\rvx_t)_i p(\rvx_t,t)\right]}{\partial x_i}  
   +\sum_{i=1}^d\sum_{j=1}^d \frac{\partial^2 }{\partial x_i \partial x_j } 
   [(\mG(t) \mG(t)^\top)_{i,j}p(\rvx_t,t)],
   \label{eq:fpk_pde}
  \end{align}
 where $(\vf(t)\rvx_t)_i$ is the $i$-th element of the vector 
 $\vf(t)\rvx_t$, and $(\mG(t) \mG(t)^\top)_{i,j}$ is 
 the element in the $i$-th row and $j$-th column of the matrix
 $\mG(t) \mG(t)^\top$. 
 Thus in this case we can derive that 
 $p(\mathbf{x}(t)|\mathbf{x}(0))$ 
 is Gaussian with mean $\vmu_t$ and variance
 $\mSigma_t$ satisfy the following ordinary differential 
 equations (ODEs)~\cite{sarkka2019applied}
\begin{align} 
  \frac{d\vmu_t}{dt} &= \vf(t) \vmu_t, \label{app:ode_mean} \\
  \frac{d\mSigma_t}{dt} &=\vf(t) \mSigma_t + \left[ \vf(t) \mSigma_t \right]^\top + 
  \mG(t) \mG(t)^\top . \label{app:ode_var}
\end{align}
This array of ODE is easy to solve, and we have
\begin{align} 
   \vmu_t &= \exp\left[\int_0^t\vf(\tau)d\tau \right] \vmu_{0}, \\
   \mSigma_t &= \exp\left[ \int_0^t\vf(\tau)d\tau \right] \mSigma_0 
   \exp\left[ \int_0^t\vf(\tau)d\tau \right]^\top  +  \int_{0}^t  \exp\left[ \int_s^t\vf(\tau)d\tau \right] \mG(s) \mG(s)^\top 
   \exp\left[ \int_s^t\vf(\tau)d\tau  \right]^\top ds
   \label{eq:app_mean_var}
 \end{align}
 where
 \begin{align} 
   \int_0^t\vf(\tau)d\tau =  
   t \begin{pmatrix} 0 &1&0\\ 
     -1 &0&\sqrt{10}\\ 
     0 & -\sqrt{10} & -6 \end{pmatrix} \otimes \mI_d.
 \end{align}
   Let $\mF =  \begin{pmatrix} 0 &1&0\\ 
      -1 &0&\sqrt{10}\\ 
      0 & -\sqrt{10} & -6 \end{pmatrix}  $, then the eigenvalues 
      of  $\mF$ are  $-3$,  $-2$, and
         $-1$. By the Putzer’s spectral 
   formula~\cite{putzer1966avoiding} and Lemma~\ref{lem:ekp} 
   we have
   \begin{align}
     \label{eq:F_put}
     \exp\left[\int_0^t\vf(\tau)d\tau \right]=\exp(t\mF\otimes \mI_d)
     =\left[r_1(t)\mP_1+r_2(t)\mP_2+r_3(t)\mP_3\right]\otimes \mI_d,
   \end{align}
   where
   \begin{align}
      \mP_1=\mI, \quad \mP_2=(\mF+3\mI)\mP_1, \quad \mP_3=(\mF+2\mI)\mP_2, 
    \end{align}
      and 
    \begin{align}
      &r'_1(t)=-3r_1(t), \quad r_1(0)=1, \\
      &r'_2(t)=-2r_2(t)+r_1(t), \quad r_2(0)=0,\\
      &r'_3(t)=-r_3(t)+r_2(t), \quad r_3(0)=0. 
    \end{align}
By integration of these ODEs, therefore
\begin{equation} 
  \begin{split}
  &\mP_1=\mI, \quad
  \mP_2=\begin{pmatrix} 3 &1&0\\ 
    -1 &3&\sqrt{10}\\ 
    0 & -\sqrt{10} & -3 \end{pmatrix},  \quad
    \mP_3=\begin{pmatrix} 5&5&\sqrt{10}\\ 
    -5 &-5 & -\sqrt{10}\\ 
    \sqrt{10} & \sqrt{10} & 2 \end{pmatrix},   \label{eq:mP}
  \end{split}
\end{equation} 
and
\begin{equation} 
   \begin{split}
   &r_1(t)=\exp\left(-3t\right),   \quad 
   r_2(t)=\exp\left(-2t\right) - \exp\left(-3t\right),\\
   &r_3(t)=\frac{1}{2}\exp\left(-t\right)  + \frac{1}{2}\exp\left(-3t\right) - \exp\left(-2t\right).  \label{eq:r}
   \end{split}
 \end{equation} 
Plugging~(\ref{eq:mP}) and~(\ref{eq:r}) into~(\ref{eq:F_put}), and 
by direct calculation
we obtain the elements of $\exp(t\mF)$ are
\begin{align}
&f_{11}(t)= \frac{5}{2}\exp(-t) -2\exp(-2t) + \frac{1}{2}\exp(-3t), \\
&f_{12}(t)= \frac{5}{2}\exp(-t) - 4\exp(-2t) + \frac{3}{2}\exp(-3t), \\
&f_{13}(t)= \sqrt{10}\left[\frac{1}{2}\exp\left(-t\right)  - \exp\left(-2t\right) + \frac{1}{2}\exp\left(-3t\right) \right], \\
&f_{21}(t)= - \frac{5}{2}\exp(-t) + 4\exp(-2t) - \frac{3}{2}\exp(-3t), \\
&f_{22}(t)= - \frac{5}{2}\exp(-t) + 8\exp(-2t) - \frac{9}{2}\exp(-3t), \\ 
&f_{23}(t)= \sqrt{10}\left[- \frac{1}{2}\exp\left(-t\right)  + 2\exp\left(-2t\right) - \frac{3}{2}\exp\left(-3t\right) \right], \\
&f_{31}(t)= \sqrt{10}\left[\frac{1}{2}\exp\left(-t\right)  - \exp\left(-2t\right) + \frac{1}{2}\exp\left(-3t\right) \right], \\
&f_{32}(t)= \sqrt{10}\left[ \frac{1}{2}\exp\left(-t\right)  - 2\exp\left(-2t\right) + \frac{3}{2}\exp\left(-3t\right) \right], \\
&f_{33}(t)= \exp(-t) -5\exp(-2t) + 5\exp(-3t). 
\end{align}
Let  $\vmu_{0} = (\rvq_{0}, \rvp_{0}, \rvs_{0})^\top$, thus the 
mean is 
\begin{align}
  \vmu_t = \begin{pmatrix} f_{11}(t)\rvq_{0} + f_{12}(t)\rvp_{0} + f_{13}(t)\rvs_{0} \\ 
    f_{21}(t)\rvq_{0} + f_{22}(t)\rvp_{0} + f_{23}(t)\rvs_{0} \\
    f_{31}(t)\rvq_{0} + f_{32}(t)\rvp_{0} + f_{33}(t)\rvs_{0} 
  \end{pmatrix}.  \label{eq:mean}
\end{align}
For the variance, let  
$\mSigma_0 = \mathrm{diag}(\Sigma_0^{qq}, \Sigma_0^{pp}, \Sigma_0^{ss}) \otimes \mI_d$, plugging the 
formula of $\exp(t\mF)$ (Eq.~(\ref{eq:F_put})) into 
$\mSigma_t$ (Eq.~(\ref{eq:app_mean_var})), implies that 
$\mSigma_t$ equals
\begin{align} 
& \exp\left[ \int_0^t\vf(\tau)d\tau \right] \mSigma_0 
  \exp\left[ \int_0^t\vf(\tau)d\tau \right]^\top   +  \int_{0}^t  \exp\left[ \int_s^t\vf(\tau)d\tau \right] \mG(s) \mG(s)^\top 
  \exp\left[ \int_s^t\vf(\tau)d\tau  \right]^\top ds \\
  &= \left[\begin{pmatrix} f_{11}(t) &f_{12}(t)&f_{13}(t)\\ 
    f_{21}(t) &f_{22}(t)&f_{23}(t)\\ 
    f_{31}(t) &f_{32}(t)&f_{33}(t)  \end{pmatrix}  
    \begin{pmatrix} \Sigma_0^{qq} &0&0\\ 
      0 &\Sigma_0^{pp}&0\\ 
      0 &0&\Sigma_0^{ss}  \end{pmatrix}  
      \begin{pmatrix} f_{11}(t) &f_{12}(t)&f_{13}(t)\\ 
        f_{21}(t) &f_{22}(t)&f_{23}(t)\\ 
        f_{31}(t) &f_{32}(t)&f_{33}(t)  \end{pmatrix}^\top\right] \otimes \mI_d \\
&+\int_{0}^t \left[ \begin{pmatrix} f_{11}(t-s) &f_{12}(t-s)&f_{13}(t-s)\\ 
  f_{21}(t-s) &f_{22}(t-s)&f_{23}(t-s)\\ 
  f_{31}(t-s) &f_{32}(t-s)&f_{33}(t-s)  \end{pmatrix} 
  \begin{pmatrix} 0 &0&0\\ 
  0 &0&0\\ 
 0 &0&12 L^{-1}  \end{pmatrix} 
 \begin{pmatrix} f_{11}(t-s) &f_{12}(t-s)&f_{13}(t-s)\\ 
  f_{21}(t-s) &f_{22}(t-s)&f_{23}(t-s)\\ 
  f_{31}(t-s) &f_{32}(t-s)&f_{33}(t-s)  \end{pmatrix} ^\top\right] \otimes \mI_d ds \\
  &= \begin{pmatrix} \sum_{j=1}^3 f_{1j}^2(t)\Sigma_0^{jj} & \sum_{j=1}^3 f_{1j}(t)f_{2j}(t)\Sigma_0^{jj} & \sum_{j=1}^3 f_{1j}(t)f_{3j}(t)\Sigma_0^{jj} \\ 
    \sum_{j=1}^3 f_{2j}(t)f_{1j}(t)\Sigma_0^{jj} & \sum_{j=1}^3 f_{2j}^2(t)\Sigma_0^{jj} & \sum_{j=1}^3 f_{2j}(t)f_{3j}(t)\Sigma_0^{jj} \\ 
    \sum_{j=1}^3 f_{3j}(t)f_{1j}(t)\Sigma_0^{jj} & \sum_{j=1}^3 f_{3j}(t)f_{2j}(t)\Sigma_0^{jj} & \sum_{j=1}^3 f_{3j}^2(t)\Sigma_0^{jj}  \end{pmatrix}   \otimes \mI_d \\
&+ 12 L^{-1} \int_{0}^t  \begin{pmatrix}  f_{13}^2(t-s) &  f_{13}(t-s)f_{23}(t-s) &  f_{13}(t-s)f_{33}(t-s) \\ 
   f_{23}(t-s)f_{13}(t-s) &  f_{23}^2(t-s) &  f_{23}(t-s)f_{33}(t-s) \\ 
 f_{33}(t-s)f_{13}(t-s) & f_{33}(t-s)f_{23}(t-s) &  f_{33}^2(t-s)  \end{pmatrix}  ds  \otimes \mI_d
\end{align}
where we let $1\leftrightarrow q, 2\leftrightarrow p, 3\leftrightarrow s$ in $\Sigma_0^{jj}$ to simplify notation.
On the other hand we have
\begin{align}
\mSigma_t = \Sigma_t  \otimes \mI_d, \quad \Sigma_t  = 
\begin{pmatrix} \Sigma_t^{qq} & \Sigma_t^{qp} & \Sigma_t^{qs}\\ 
  \Sigma_t^{qp} & \Sigma_t^{pp} & \Sigma_t^{ps}\\ 
  \Sigma_t^{qs} & \Sigma_t^{ps} & \Sigma_t^{ss}  \end{pmatrix},
\end{align}
which clearly implies that
\begin{align}
  &\Sigma_t^{qq} = \sum_{j=1}^3 f_{1j}^2(t)\Sigma_0^{jj} +12 L^{-1} \int_{0}^t f_{13}^2(t-s) ds , \\
  &\Sigma_t^{qp}=  \sum_{j=1}^3 f_{1j}(t)f_{2j}(t)\Sigma_0^{jj} +  12 L^{-1} \int_{0}^t  f_{13}(t-s)f_{23}(t-s) ds, \\
  &\Sigma_t^{qs}= \sum_{j=1}^3 f_{1j}(t)f_{3j}(t)\Sigma_0^{jj} + 12 L^{-1}  \int_{0}^t f_{13}(t-s)f_{33}(t-s)  ds, \\
  &\Sigma_t^{pp}= \sum_{j=1}^3 f_{2j}^2(t)\Sigma_0^{jj} + 12 L^{-1}  \int_{0}^t  f_{23}^2(t-s) ds, \\
  &\Sigma_t^{ps}= \sum_{j=1}^3 f_{2j}(t)f_{3j}(t)\Sigma_0^{jj}  + 12 L^{-1}  \int_{0}^t  f_{23}(t-s)f_{33}(t-s) ds, \\ 
  &\Sigma_t^{ss}= \sum_{j=1}^3 f_{3j}^2(t)\Sigma_0^{jj}  + 12 L^{-1}  \int_{0}^t  f_{33}^2(t-s)ds,
  \end{align}
where
\begin{align}
   &f_{13}^2(t)= f_{31}^2(t)=\frac{5}{2}\exp(-2t) - 10\exp(-3t) + 15\exp(-4t)  - 10\exp(-5t)+\frac{5}{2}\exp(-6t), \\
   &f_{23}^2(t)= f_{32}^2(t)= \frac{5}{2}\exp(-2t) - 20\exp(-3t) + 55\exp(-4t)  - 60\exp(-5t)+\frac{45}{2}\exp(-6t), \\
   &f_{33}^2(t)=  \exp(-2t) - 10 \exp(-3t) + 35\exp(-4t)  - 50\exp(-5t)+25\exp(-6t), \\
  % \end{align}
  % \begin{align}
   &f_{13}(t)f_{23}(t)=-\frac{5}{2}\exp(-2t) +15\exp(-3t) -30\exp(-4t)  + 25\exp(-5t)-\frac{15}{2}\exp(-6t), \\
   &f_{13}(t)f_{33}(t)=\frac{\sqrt{10}}{2}\exp(-2t) -\frac{7\sqrt{10}}{2}\exp(-3t) + 8\sqrt{10}\exp(-4t)  - \frac{15\sqrt{10}}{2}\exp(-5t)+\frac{5\sqrt{10}}{2}\exp(-6t), \\
   &f_{23}(t)f_{33}(t)= \sqrt{10}\left[-\frac{1}{2}\exp(-2t) 
   +\frac{9}{2}\exp(-3t) - 14\exp(-4t)  
   + \frac{35}{2}\exp(-5t)-\frac{15}{2}\exp(-6t)\right], \\
%  \end{align}
%    %
%    \begin{align}
     &f_{11}^2(t)= \frac{25}{4}\exp(-2t) - 10\exp(-3t) + \frac{13}{2}\exp(-4t) - 2\exp(-5t)+\frac{1}{4}\exp(-6t), \\
     &f_{12}^2(t)= f_{21}^2(t)=\frac{25}{4}\exp(-2t) - 20\exp(-3t) + \frac{47}{2}\exp(-4t) - 12\exp(-5t)+\frac{9}{4}\exp(-6t),\\
     &f_{22}^2(t)= \frac{25}{4}\exp(-2t) - 40\exp(-3t) + 86.5\exp(-4t)  - 72\exp(-5t)+\frac{81}{4}\exp(-6t), \\
  %  \end{align}
  %  \begin{align}
     &f_{21}(t)f_{31}(t)=-\frac{5\sqrt{10}}{4}\exp(-2t) +\frac{9\sqrt{10}}{2}\exp(-3t) - 6\sqrt{10}\exp(-4t)  + \frac{7\sqrt{10}}{2}\exp(-5t)-\frac{3\sqrt{10}}{4}\exp(-6t), \\
     &f_{22}(t)f_{32}(t)=\sqrt{10}\left[-\frac{5}{4}\exp(-2t) 
     + 9\exp(-3t) -22\exp(-4t)  
     + 21\exp(-5t)-\frac{27}{4}\exp(-6t)\right], \\
     &f_{11}(t)f_{31}(t)= \frac{5\sqrt{10}}{4}\exp(-2t) -\frac{7\sqrt{10}}{2}\exp(-3t) + \frac{7\sqrt{10}}{2}\exp(-4t)  - \frac{3\sqrt{10}}{2}\exp(-5t)+\frac{\sqrt{10}}{4}\exp(-6t), \\ 
    % \end{align}
    % \begin{align}
     &f_{12}(t)f_{32}(t)=\frac{5\sqrt{10}}{4}\exp(-2t) -7\sqrt{10}\exp(-3t) + \frac{25\sqrt{10}}{2}\exp(-4t)  - 9\sqrt{10}\exp(-5t)+\frac{9\sqrt{10}}{4}\exp(-6t), \\
     &f_{11}(t)f_{21}(t)=-\frac{25}{4}\exp(-2t) + 15\exp(-3t) -13\exp(-4t)  +5\exp(-5t)-\frac{3}{4}\exp(-6t), \\
     &f_{12}(t)f_{22}(t)=-\frac{25}{4}\exp(-2t) + 30\exp(-3t) -47\exp(-4t)  +30\exp(-5t)-\frac{27}{4}\exp(-6t).
\end{align}
Due to the fact  $\int_{0}^t \exp(-a(t-s)) ds = \frac{1-\exp(-at)}{a}$, 
so we 
can deduce that  $\lim_{t\rightarrow\infty} \int_{0}^t \exp(-a(t-s)) ds =  \frac{1}{a}$.
   Then by direct computation we have
\begin{align}
  \lim_{t\rightarrow\infty} \Sigma^{qq}_t &=2\xi L^{-1}\frac{1}{12}=L^{-1}, \\
  \lim_{t\rightarrow\infty} \Sigma^{pp}_t &=2\xi L^{-1}\frac{1}{12}=L^{-1}, \\
  \lim_{t\rightarrow\infty} \Sigma^{ss}_t &=2\xi L^{-1}\frac{1}{12}=L^{-1}, \\
  \lim_{t\rightarrow\infty} \Sigma^{qp}_t &=0, \\
  \lim_{t\rightarrow\infty} \Sigma^{qs}_t &=0, \\
  \lim_{t\rightarrow\infty} \Sigma^{ps}_t &=0, \\
  \lim_{t\rightarrow\infty} \vmu_t & = \boldsymbol{0}_{3d},
\end{align}
which establishes the prior probability 
 $p_{\infty}(\rvx)=\mathcal{N}(\rvq;\bm{0}_d,L^{-1}\mI_d)\,\mathcal{N}(\rvp;\bm{0}_d,L^{-1}\mI_d)\,\mathcal{N}(\rvs;\bm{0}_d,L^{-1}\mI_d)$.
\subsection{The Objective}

Let  $p(\rvx_0, 0)$ denotes the distribution of 
target data (image, video, or audio etc. with initial 
velocity and acceleration) that we 
want to 
generate, and $p(\rvx_0, 0)$ can be diffused 
through $p(\rvx_t, t)$ (sometimes it is abbreviated as $p_t$ 
for brevity) into $p(\rvx_T, T)$ by 
HOLD~(Eq.~(\ref{eq:app_hold})).
The target of generation is to build a model to 
reverse this process, that is use time-reverse HOLD with 
Eq.~(\ref{eq:app_hold_matrix}) to sample 
from $q(\rvx_T, T)$ (approximation of the prior probability) 
and denoise it through $q(\rvx_t, t)$ into  $q(\rvx_0, 0)$.
Thus our goal (or objective) is to make the  difference between 
$q(\rvx_0, 0)$ and $p(\rvx_0, 0)$ as small as possible.
Kullback–Leibler (KL) divergence is an effective measure to  
fulfil by
\begin{equation}
  \begin{split}
      \infdiv{p_0}{q_0} &= \infdiv{p_0}{q_0} - \infdiv{p_T}{q_T} + \infdiv{p_T}{q_T} \\
      &= - \int_0^T \frac{\partial\infdiv{p_t}{q_t}}{\partial t}dt + \infdiv{p_T}{q_T} \label{eq:kl_objective}.
  \end{split}
\end{equation}
If $T$ is large,  $\infdiv{p_T}{q_T}$ is small enough to 
be ignored, so the most important 
thing to calculate an objective is to know the value of $\int_0^T \frac{\partial\infdiv{p_t}{q_t}}{\partial t}dt$,
whose deduction needs $\frac{\partial p(\rvx_t, t)}{\partial t}$, which can be obtained by 
the Fokker--Planck equation associated with our diffusion HOLD
\begin{equation}
      \begin{split}
          \frac{\partial p(\rvx_t, t)}{\partial t} &= \nabla_{\rvx} \cdot 
          \left[\tfrac{1}{2} \left(\mG(t) \mG(t)^\top \otimes \mI_d\right) 
          \nabla_{\rvx} p(\rvx_t, t) - p(\rvx_t, t) (\vf(t) \otimes \mI_d) \rvx_t \right] \\
          &= \nabla_{\rvx_t} \cdot \left[\vh_p(\rvx_t, t) p(\rvx_t, t)\right], 
          \quad \vh_p(\rvx_t, t) 
          \coloneqq \tfrac{1}{2} \left(\mG(t) \mG(t)^\top  \otimes \mI_d\right) 
          \nabla_{\rvx} \log p(\rvx_t, t) - (\vf(t) \otimes \mI_d) \rvx_t,
      \end{split}
  \end{equation}
  where $\nabla_{\rvx_t} \cdot$ is divergence, and $\nabla_{\rvx_t}$ is gradient.
Then with Gauss's theorem, we have
\begin{equation}
\begin{split}
  & \quad \frac{\partial\infdiv{p_t}{q_t}}{\partial t} = \frac{\partial}{\partial t} \int p(\rvx_t, t) \log \frac{p(\rvx_t, t)}{q(\rvx_t, t)} \, d\rvx_t \\
   &=  \int \frac{\partial p(\rvx_t, t)}{\partial t} \log \frac{p(\rvx_t, t)}{q(\rvx_t, t)} 
   + p(\rvx_t, t) \frac{\partial \log \frac{p(\rvx_t, t)}{q(\rvx_t, t)}}{\partial t} \, d\rvx_t \\
   &=  \int \nabla_{\rvx_t} \cdot \left[\vh_p(\rvx_t, t) p(\rvx_t, t)\right] \left[ \log {p(\rvx_t, t)} - \log {q(\rvx_t, t)} \right]
   +  \frac{\partial p(\rvx_t, t)}{\partial t} - \frac{p(\rvx_t, t)}{q(\rvx_t, t)}\frac{\partial q(\rvx_t, t)}{\partial t} \, d\rvx_t \\
   &=  \int \left[\vh_p(\rvx_t, t) p(\rvx_t, t)\right] \nabla\left[ \log {p(\rvx_t, t)} - \log {q(\rvx_t, t)} \right]
   +  \nabla_{\rvx_t} \cdot \left[\vh_p(\rvx_t, t) p(\rvx_t, t)\right]  - \frac{p(\rvx_t, t)}{q(\rvx_t, t)}
   \nabla_{\rvx_t} \cdot \left[\vh_q(\rvx_t, t) q(\rvx_t, t)\right]  \, d\rvx_t \\
  &=  \int -\left[\vh_p(\rvx_t, t) p(\rvx_t, t)\right]^\top \nabla\left[ \log {p(\rvx_t, t)} - \log {q(\rvx_t, t)} \right]
  + 
  \left[\vh_q(\rvx_t, t) q(\rvx_t, t)\right]^\top \nabla_{\rvx_t} \frac{p(\rvx_t, t)}{q(\rvx_t, t)} \, d\rvx_t \\
  &=  \int -\left[\vh_p(\rvx_t, t) p(\rvx_t, t)\right]^\top \nabla\left[ \log {p(\rvx_t, t)} - \log {q(\rvx_t, t)} \right] + 
  \left[\vh_q(\rvx_t, t) q(\rvx_t, t)\right]^\top  \frac{q(\rvx_t, t)\nabla_{\rvx_t} p(\rvx_t, t) - p(\rvx_t, t)\nabla_{\rvx_t} q(\rvx_t, t)}{q^2(\rvx_t, t)} \, d\rvx_t \\
  &= - \int p(\rvx_t, t) \left[\vh_p(\rvx_t, t) -  \vh_q(\rvx_t, t)\right]^\top \left[\nabla_{\rvx_t} \log p(\rvx_t, t) - \nabla_{\rvx_t} \log q(\rvx_t, t)\right] \, d\rvx_t \\
    &= - \frac{1}{2} \int p(\rvx_t, t) \left[\nabla_{\rvx_t} \log p(\rvx_t, t) - \nabla_{\rvx_t} \log q(\rvx_t, t)\right]^\top \left(G(t) G(t)^\top \otimes \mI_d\right) 
    [\nabla_{\rvx_t} \log p(\rvx_t, t)  - \nabla_{\rvx_t} \log q(\rvx_t, t) ] \, d\rvx_t \\
    &= - 6L^{-1}  \int p(\rvx_t, t) \norm{\nabla_{\rvs_t} \log p(\rvx_t, t) - \nabla_{\rvs_t} \log q(\rvx_t, t)}_2^2 \, d\rvx_t.
\end{split}
\end{equation}
Thus the objective is 
\begin{equation}
\begin{split}
    \infdiv{p_0}{q_0} &= \E_{t \sim \gU[0, T], \rvx_t \sim p(\rvx, t)} \left[6L^{-1} \norm{\nabla_{\rvs_t} \log p(\rvx_t, t) - \nabla_{\rvs_t} \log q(\rvx_t, t)}_2^2 \right] + \infdiv{p_T}{q_T} \\
    &\approx \E_{t \sim \gU[0, T], \rvx_t \sim p(\rvx, t)} \left[6L^{-1} \norm{\nabla_{\rvs_t} \log p(\rvx_t, t) - \nabla_{\rvs_t} \log q(\rvx_t, t)}_2^2 \right] \\
    &= \E_{t \sim \gU[0, T], \rvx_t \sim p(\rvx, t)} \left[\lambda(t) \norm{\nabla_{\rvs_t} \log p(\rvx_t, t) - \nabla_{\rvs_t} \log q(\rvx_t, t)}_2^2 \right],
\end{split}
\label{eq:app_hold_sm_obj}
\end{equation}
where a more general objective function can be obtained by 
replacing $6L^{-1}$ with an arbitrary function $\lambda(t)$.
\subsection{Denoising Score Matching and Hybrid Score Matching}
In this work, a neural network is used to estimate the score 
$\nabla_{\rvs_t} \log p(\rvx_t, t)$.
Let $\mathfrak{S}_\vtheta(\rvx_t, t)$ denotes this score model. 
Substituting $\mathfrak{S}_\vtheta(\rvx_t, t)$ 
for $\nabla_{\rvs_t} \log q(\rvx_t, t)$ in 
Eq.~(\ref{eq:app_hold_sm_obj}) gives the score matching (SM)
loss
\begin{equation} \label{eq:app_sm_loss}
  \gL_{\mathrm{SM}} \coloneqq \E_{t \sim \gU[0, T]} \left[\lambda(t) \E_{\rvx_t \sim p(\rvx, t)} 
  [\norm{\nabla_{\rvs_t} \log p(\rvx_t, t) - \mathfrak{S}_\vtheta(\rvx_t, t)}_2^2] \right].
\end{equation}
 In order to achieve high-precision score estimation on a low-dimensional data manifold, 
 the equivalent denoising score matching (DSM) 
 loss~\cite{vincent2011connection,song2019generative}  
 \begin{equation} \label{eq:app_dsm_loss}
\gL_{\mathrm{DSM}} \coloneqq \E_{t \sim \gU[0, T]} \left[\lambda(t)  
\E_{\rvx_0 \sim p(\rvx_0, 0), \rvx_t \sim p(\rvx_t \mid \rvx_0, t)} 
\norm{\nabla_{\rvs_t} \log p(\rvx_t \mid \rvx_0, t ) - \mathfrak{S}_\vtheta(\rvx_t, t)}_2^2 \right]
\end{equation}
apply a strategy by contaminating the data with a very small-scale noise 
to make the signal  spread throughout the 
entire ambient Euclidean space instead of being limited to a  low-dimensional manifold.
By marginalizing over the entire initial auxiliary signal, 
Tim~\cite{dockhorn2021score} propose the hybrid 
score matching (HSM)
\begin{equation} \label{eq:app_hsm_loss}
      \gL_{\mathrm{HSM}} \coloneqq \E_{t \sim \gU[0, T]} \left[\lambda(t)   
      \E_{\rvq_0 \sim p(\rvq_0), \rvx_t \sim p(\rvx_t \mid \rvq_0, t)} 
      \norm{\nabla_{\rvs_t} \log p(\rvx_t \mid \rvq_0, t) - \mathfrak{S}_\vtheta(\rvx_t, t)}_2^2 \right].
\end{equation}

In fact, SM~(Eq.~(\ref{eq:app_sm_loss})), 
DSM~(Eq.~(\ref{eq:app_dsm_loss})) and 
HSM~(Eq.~(\ref{eq:app_hsm_loss})) are equivalent.
The equivalence between SM
and DSM can be derived as following
 \begin{equation*} 
   \begin{split} \label{app:sm_equal_dsm}
       &\gL_{\mathrm{SM}} = \E_{t \sim \gU[0, T]}\lambda(t) \left[\E_{\rvx_t \sim p(\rvx_t, t)} 
       \norm{\mathfrak{S}_\vtheta(\rvx_t, t)}_2^2 - 
       2 \E_{\rvx_t \sim p(\rvx_t, t)} \left \langle \nabla_{\rvs_t} \log p(\rvx_t, t), \mathfrak{S}_\vtheta(\rvx_t, t) \right \rangle + C\right] \\
       &=  \E_{t \sim \gU[0, T]}\lambda(t) \left[\E_{\rvx_t \sim p(\rvx_t, t)} 
       \norm{\mathfrak{S}_\vtheta(\rvx_t, t)}_2^2 - \int_{\rvx_t} p(\rvx_t, t) \left \langle \nabla_{\rvs_t} \log p(\rvx_t, t), \mathfrak{S}_\vtheta(\rvx_t, t) \right \rangle \,d\rvx_t + C\right] \\
       &= \E_{t \sim \gU[0, T]} \lambda(t)\left[\E_{\rvx_t \sim p(\rvx_t, t)} 
       \norm{\mathfrak{S}_\vtheta(\rvx_t, t)}_2^2 - \int_{\rvx_t} \left \langle \nabla_{\rvs_t} p(\rvx_t, t), \mathfrak{S}_\vtheta(\rvx_t, t) \right \rangle \,d\rvx_t  + C\right] \\
       &= \E_{t \sim \gU[0, T]} \lambda(t) \left[\E_{\rvx_t \sim p(\rvx_t, t)} 
       \norm{\mathfrak{S}_\vtheta(\rvx_t, t)}_2^2 -  \int_{\rvx_t} \left \langle \nabla_{\rvs_t} \int_{\rvx_0} p(\rvx_t \mid \rvx_0, t) p(\rvx_0, 0) \, d\rvx_0, \mathfrak{S}_\vtheta(\rvx_t, t) \right \rangle \,d\rvx_t  + C\right]\\
       &= \E_{t \sim \gU[0, T]} \lambda(t) \left[\E_{\rvx_t \sim p(\rvx_t, t)} 
       \norm{\mathfrak{S}_\vtheta(\rvx_t, t)}_2^2 -  \int_{\rvx_t} \left \langle \int_{\rvx_0} p(\rvx_t \mid \rvx_0, t) p(\rvx_0, 0) \nabla_{\rvs_t} \log p(\rvx_t \mid \rvx_0, t) \, d\rvx_0, \mathfrak{S}_\vtheta(\rvx_t, t) \right \rangle \,d\rvx_t  + C\right]\\
       &= \E_{t \sim \gU[0, T]}\lambda(t) \left[\E_{\rvx_t \sim p(\rvx_t, t)} 
       \norm{\mathfrak{S}_\vtheta(\rvx_t, t)}_2^2 -  \int_{\rvx_t} \int_{\rvx_0} p(\rvx_t \mid \rvx_0, t) p(\rvx_0, 0) \left \langle \nabla_{\rvs_t} \log p(\rvx_t \mid \rvx_0, t), \mathfrak{S}_\vtheta(\rvx_t, t) \right \rangle \, d\rvx_0 \,d\rvx_t  + C\right] \\
       &= \E_{t \sim \gU[0, T]}\lambda(t)\left[\E_{\rvx_t \sim p(\rvx_t, t)} 
       \norm{\mathfrak{S}_\vtheta(\rvx_t, t)}_2^2 -  \E_{\rvx_0 \sim p(\rvx_0, 0), \rvx_t \sim p(\rvx_t \mid \rvx_0)} \left[\left \langle \nabla_{\rvs_t} \log p(\rvx_t \mid \rvx_0, t), \mathfrak{S}_\vtheta(\rvx_t, t) \right \rangle \right]  + C\right]\\
       &=\E_{t \sim \gU[0, T]} \left[\lambda(t)  
       \E_{\rvx_0 \sim p(\rvx_0), \rvx_t \sim p_t(\rvx_t \mid \rvx_0)} 
       \norm{\nabla_{\rvs_t} \log p_t(\rvx_t \mid \rvx_0) - \mathfrak{S}_\vtheta(\rvx_t, t)}_2^2 \right] +C = \gL_{\mathrm{DSM}} +C,
   \end{split}
   \end{equation*}
where in the formulas corresponding to different 
``$=$''s, $C$ may be different, 
but represents a term that does not depend 
on $\vtheta$.

The equivalence between 
SM and HSM can be 
deducted as 
\begin{equation*} 
  \begin{split} \label{app:sm_equal_hsm}
      &\gL_{\mathrm{SM}} = \E_{t \sim \gU[0, T]}\lambda(t) \left[\E_{\rvx_t \sim p(\rvx_t, t)} 
      \norm{\mathfrak{S}_\vtheta(\rvx_t, t)}_2^2 - 
      2 \E_{\rvx_t \sim p(\rvx_t, t)} \left \langle \nabla_{\rvs_t} \log p(\rvx_t, t), \mathfrak{S}_\vtheta(\rvx_t, t) \right \rangle + C\right] \\
      &= \E_{t \sim \gU[0, T]}\lambda(t) \left[\E_{\rvx_t \sim p(\rvx_t, t)} 
      \norm{\mathfrak{S}_\vtheta(\rvx_t, t)}_2^2 - \int_{\rvx_t} p(\rvx_t, t) \left \langle \nabla_{\rvs_t} \log p(\rvx_t, t), \mathfrak{S}_\vtheta(\rvx_t, t) \right \rangle \,d\rvx_t + C\right]\\
          &= \E_{t \sim \gU[0, T]}\lambda(t) \left[\E_{\rvx_t \sim p(\rvx_t, t)} 
          \norm{\mathfrak{S}_\vtheta(\rvx_t, t)}_2^2 - \int_{\rvx_t} \left \langle \nabla_{\rvs_t} p(\rvx_t, t), \mathfrak{S}_\vtheta(\rvx_t, t) \right \rangle \,d\rvx_t + C\right]\\
          &= \E_{t \sim \gU[0, T]}\lambda(t) \left[\E_{\rvx_t \sim p(\rvx_t, t)} 
          \norm{\mathfrak{S}_\vtheta(\rvx_t, t)}_2^2 - \int_{\rvx_t} \left \langle \nabla_{\rvs_t} \int_{\rvq_0} p(\rvx_t \mid \rvq_0, t) p(\rvq_0, 0) \, d\rvq_0, \mathfrak{S}_\vtheta(\rvx_t, t) \right \rangle \,d\rvx_t + C\right]\\
          &= \E_{t \sim \gU[0, T]}\lambda(t) \left[\E_{\rvx_t \sim p(\rvx_t, t)} 
          \norm{\mathfrak{S}_\vtheta(\rvx_t, t)}_2^2 - \int_{\rvx_t} \left \langle \int_{\rvq_0} p(\rvx_t \mid \rvq_0, t) p(\rvq_0, 0) \nabla_{\rvs_t} \log p(\rvx_t \mid \rvq_0, t) \, d\rvq_0, \mathfrak{S}_\vtheta(\rvx_t, t) \right \rangle \,d\rvx_t + C\right]\\
          &=\E_{t \sim \gU[0, T]}\lambda(t) \left[\E_{\rvx_t \sim p(\rvx_t, t)} 
          \norm{\mathfrak{S}_\vtheta(\rvx_t, t)}_2^2 - \int_{\rvx_t} \int_{\rvq_0} p(\rvx_t \mid \rvq_0, t) p(\rvq_0, 0) \left \langle \nabla_{\rvs_t} \log p(\rvx_t \mid \rvq_0, t), \mathfrak{S}_\vtheta(\rvx_t, t) \right \rangle \, d\rvq_0 \,d\rvx_t + C\right]\\
          &= \E_{t \sim \gU[0, T]}\lambda(t) \left[\E_{\rvx_t \sim p(\rvx_t, t)} 
          \norm{\mathfrak{S}_\vtheta(\rvx_t, t)}_2^2 - \E_{\rvq_0 \sim p(\rvq_0, 0), \rvx_t \sim p(\rvx_t \mid \rvq_0)} \left[\left \langle \nabla_{\rvs_t} \log p(\rvx_t \mid \rvq_0, t), \mathfrak{S}_\vtheta(\rvx_t, t) \right \rangle \right] + C\right]\\
     &=\E_{t \sim \gU[0, T]}\lambda(t) \left[  
     \E_{\rvq_0 \sim p(\rvq_0), \rvx_t \sim p(\rvx_t \mid \rvq_0, t)} 
     \norm{\nabla_{\rvs_t} \log p(\rvx_t \mid \rvq_0, t) - \mathfrak{S}_\vtheta(\rvx_t, t)}_2^2 \right]+C = \gL_{\mathrm{HSM}} +C.
  \end{split}
  \end{equation*}

\subsection{Block Coordinate Score Matching}
\label{sec:app_bcsm}

For general high-order dynamics HOLD, there are 
multiple variables (several blocks of coordinates), e.g. 
for third-order Langevin dynamic system  only one 
block $\rvq$ to represent the variables of interest and other two 
blocks $\rvp$ and $\rvs$ are introduced just to allow the two 
Hamiltonian dynamics to operate. There is no obvious reason that 
we need to denoise all blocks. So in this paper we propose
block coordinate score matching (BCSM)
to 
denoise only initial distribution  for  part $\rvb_0$ of 
all blocks $\rvx_0$, and 
marginalize over the distribution $p(\rvx_0 \backslash \rvb_0)$ of
remaining variables, which results in
 \begin{equation} \label{eq:app_bcsm_loss}
  \gL_{\mathrm{BCSM}} \coloneqq \E_{t \sim \gU[0, T]} 
  \left[\lambda(t)   
  \E_{\rvb_0 \sim p(\rvb_0), \rvx_t \sim p(\rvx_t \mid \rvb_0, t)} 
  \norm{\nabla_{\rvs_t} \log p(\rvx_t \mid \rvb_0, t) - \mathfrak{S}_\vtheta(\rvx_t, t)}_2^2 \right].
\end{equation}
BCSM is a flexible objective, and the only requirement for 
$\rvb_0$ is that it must contain block
$\rvq_0$, 
which is the variable we are interested in.
BCSM is also equivalent to SM.
 Indeed we have
 \begin{equation} 
   \begin{split}
    &\gL_{\mathrm{SM}} = \E_{t \sim \gU[0, T]}\lambda(t) \left[\E_{\rvx_t \sim p(\rvx_t, t)} 
    \norm{\mathfrak{S}_\vtheta(\rvx_t, t)}_2^2 - 
    2 \E_{\rvx_t \sim p(\rvx_t, t)} \left \langle \nabla_{\rvs_t} \log p(\rvx_t, t), \mathfrak{S}_\vtheta(\rvx_t, t) \right \rangle + C\right] \\
       &= \E_{t \sim \gU[0, T]}\lambda(t) \left[\E_{\rvx_t \sim p(\rvx_t, t)} 
       \norm{\mathfrak{S}_\vtheta(\rvx_t, t)}_2^2 - \int_{\rvx_t} p(\rvx_t, t) \left \langle \nabla_{\rvs_t} \log p(\rvx_t, t), \mathfrak{S}_\vtheta(\rvx_t, t) \right \rangle \,d\rvx_t + C\right]\\
       &= \E_{t \sim \gU[0, T]}\lambda(t) \left[\E_{\rvx_t \sim p(\rvx_t, t)} 
       \norm{\mathfrak{S}_\vtheta(\rvx_t, t)}_2^2 - \int_{\rvx_t} \left \langle \nabla_{\rvs_t} p(\rvx_t, t), \mathfrak{S}_\vtheta(\rvx_t, t) \right \rangle \,d\rvx_t + C\right]\\
       &= \E_{t \sim \gU[0, T]}\lambda(t) \left[\E_{\rvx_t \sim p(\rvx_t, t)} 
       \norm{\mathfrak{S}_\vtheta(\rvx_t, t)}_2^2 - \int_{\rvx_t} \left \langle \nabla_{\rvs_t} \int_{\rvb_0} p(\rvx_t \mid \rvb_0, t) p(\rvb_0, 0) \, d\rvb_0, \mathfrak{S}_\vtheta(\rvx_t, t) \right \rangle \,d\rvx_t + C\right] \\
       &= \E_{t \sim \gU[0, T]}\lambda(t) \left[\E_{\rvx_t \sim p(\rvx_t, t)} 
       \norm{\mathfrak{S}_\vtheta(\rvx_t, t)}_2^2 - \int_{\rvx_t} \left \langle \int_{\rvb_0} p(\rvx_t \mid \rvb_0, t) p(\rvb_0, 0) \nabla_{\rvs_t} \log p(\rvx_t \mid \rvb_0, t) \, d\rvb_0, \mathfrak{S}_\vtheta(\rvx_t, t) \right \rangle \,d\rvx_t + C\right]\\
       &= \E_{t \sim \gU[0, T]}\lambda(t) \left[\E_{\rvx_t \sim p(\rvx_t, t)} 
       \norm{\mathfrak{S}_\vtheta(\rvx_t, t)}_2^2 -  \int_{\rvx_t} \int_{\rvb_0} p(\rvx_t \mid \rvb_0, t) p(\rvb_0, 0) \left \langle \nabla_{\rvs_t} \log p(\rvx_t \mid \rvb_0, t), \mathfrak{S}_\vtheta(\rvx_t, t) \right \rangle \, d\rvb_0 \,d\rvx_t+ C\right] \\
       &= \E_{t \sim \gU[0, T]}\lambda(t) \left[\E_{\rvx_t \sim p(\rvx_t, t)} 
       \norm{\mathfrak{S}_\vtheta(\rvx_t, t)}_2^2 -  \E_{\rvb_0 \sim p(\rvb_0, 0), \rvx_t \sim p(\rvx_t \mid \rvb_0)} \left[\left \langle \nabla_{\rvs_t} \log p(\rvx_t \mid \rvb_0, t), \mathfrak{S}_\vtheta(\rvx_t, t) \right \rangle \right] + C\right]
       = \gL_{\mathrm{BCSM}}  + C.
   \end{split}
   \end{equation}
   This fact and Eq.~(\ref{app:ode_mean}), Eq.~(\ref{app:ode_var}) 
   can be used to compute the gradient 
   $\nabla_{\rvx_t} \log p(\rvx_t \mid \cdot, t)$
   \begin{equation}
   \begin{split}
       \nabla_{\rvx_t} \log p(\rvx_t \mid \cdot, t) 
       = - \nabla_{\rvx_t} \tfrac{1}{2} (\rvx_t - \vmu_t) \mSigma_t^{-1} (\rvx_t - \vmu_t) 
       = - \mSigma_t^{-1} (\rvx_t - \vmu_t)
       = - \mL_t^{-\top} \mL_t^{-1} (\rvx_t - \vmu_t) 
       = - \mL_t^{-\top} \rvepsilon_{3d}, \label{eq:gradient_gaussian_transition_kernel}
   \end{split}
   \end{equation}
  and
  \begin{equation}
    \begin{split}
        \nabla_{\rvs_t} \log p_t(\rvx_t \mid \cdot) 
        = \left[\nabla_{\rvx_t} \log p_t(\rvx_t \mid \cdot)\right]_{2d:3d}
        = \left[- \mL_t^{-\top} \rvepsilon_{3d} \right]_{2d:3d}
        = -\ell_t \rvepsilon_{2d:3d},
    \end{split}
    \end{equation}
  where $\rvepsilon_{3d} \sim \gN(\bm{0}, \mI_{3d})$ 
  and $\mSigma_t = \mL_t \mL_t^\top$ is the Cholesky 
  factorization of the covariance matrix $\mSigma_t$, and
   \begin{align}
    \ell_t \coloneqq \left[\Sigma^{ss}_t - \frac{(\Sigma^{sq}_t)^2}{\Sigma^{qq}_t} - \left(\Sigma^{pp}_t - \frac{\left(\Sigma^{pq}_t\right)^2}{\Sigma^{qq}_t}\right)^{-1}
    \left(\Sigma^{sp}_t - \frac{\Sigma^{sq}_t\Sigma^{pq}_t}{\Sigma^{qq}_t} \right)^2\right]^{-0.5}.
\end{align}
Here $\rvx_t$ is sampled via reparameterization
\begin{align}
  \rvx_t = \vmu_t + \mL_t \rvepsilon  = \vmu_t + 
  \begin{pmatrix} L^{qq}_t \rvepsilon_{0:d} \\ 
    L^{pq}_t \rvepsilon_{0:d} + L^{pp}_t \rvepsilon_{d:2d} \\ 
    L^{sq}_t \rvepsilon_{0:d} + L^{sp}_t \rvepsilon_{d:2d} +  L^{ss}_t \rvepsilon_{2d:3d}
  \end{pmatrix}.
\end{align}
%. 
  Note that the structure of $\mSigma_t$ implies that $\mL_t = L_t \otimes \mI_d$, where $L_t L_t^\top$ is the Cholesky 
  factorization of $\Sigma_t$, i.e,
  $\mSigma_t = \mL_t \mL_t^\top$.
% %
% \begin{align}
%  \mSigma_t = \Sigma_t \otimes \mI_d, \quad \Sigma_t = \begin{pmatrix} \Sigma^{qq}_t &\Sigma^{qp}_t &\Sigma^{qs}_t\\ 
%    \Sigma^{pq}_t &\Sigma^{pp}_t &\Sigma^{ps}_t\\  \Sigma^{sq}_t &\Sigma^{sp}_t &\Sigma^{ss}_t \end{pmatrix}.
% \end{align}
% %
It is not difficult to obtain that
\begin{align}
  &L_t = \begin{pmatrix}
  L_t^{qq} & 0 & 0 \\ L_t^{pq} & L_t^{pp} & 0 \\ L_t^{sq} & L_t^{sp} & L_t^{ss} 
  \end{pmatrix} = \\ &
  \begin{pmatrix} \sqrt{\Sigma^{qq}_t} & 0 & 0\\ 
    \frac{\Sigma^{pq}_t}{\sqrt{\Sigma^{qq}_t}} & \sqrt{\Sigma^{pp}_t - \frac{\left(\Sigma^{pq}_t\right)^2}{\Sigma^{qq}_t}} & 0  \\
    \frac{\Sigma^{sq}_t}{\sqrt{\Sigma^{qq}_t}} & \left(\sqrt{\Sigma^{pp}_t- \frac{\left(\Sigma^{pq}_t\right)^2}{\Sigma^{qq}_t}}\right)^{-1}
    \left(\Sigma^{sp}_t - \frac{\Sigma^{sq}_t\Sigma^{pq}_t}{\Sigma^{qq}_t} \right) & 
    \sqrt{\Sigma^{ss}_t - \frac{(\Sigma^{sq}_t)^2}{\Sigma^{qq}_t} - \left(\Sigma^{pp}_t -\frac{ \left(\Sigma^{pq}_t\right)^2}{\Sigma^{qq}_t}\right)^{-1}
    \left(\Sigma^{sp}_t - \frac{\Sigma^{sq}_t\Sigma^{pq}_t}{\Sigma^{qq}_t} \right)^2} 
  \end{pmatrix}.
\end{align}
Thus we have
\begin{equation} \label{eq:app_dsm_loss_reparam}
  \gL_{\mathrm{DSM}} \coloneqq \E_{t \sim \gU[0, T]} \left[\lambda(t)  
  \E_{\rvx_0 \sim p(\rvx_0), \rvx_t \sim p_t(\rvx_t \mid \rvx_0)} 
  \norm{-\ell^{DSM}_t \rvepsilon_{2d:3d} - \mathfrak{S}_\vtheta(\rvx_t, t)}_2^2 \right]
\end{equation}
and the BCSM is 
\begin{equation} \label{eq:app_bcsm_loss_reparam}
  \gL_{\mathrm{BCSM}} \coloneqq \E_{t \sim \gU[0, T]} 
  \left[\lambda(t)   
  \E_{\rvb_0 \sim p(\rvb_0), \rvx_t \sim p(\rvx_t \mid \rvb_0, t)} 
  \norm{-\ell^{BCSM}_t \rvepsilon_{2d:3d} - \mathfrak{S}_\vtheta(\rvx_t, t)}_2^2 \right].
\end{equation}
   The intuition behind BCSM is that the distribution of  $q_0$ is complex and unknown, 
   and the distribution of $s_0$ is known, easy and fixed.

\section{Sampling with HOLD}

% If we assume $\vmu_0=(\rvq_0, \bm{0}_d, \bm{0}_d)$, and 
% $\Sigma_0^{qq}=0, \Sigma_0^{pp}=\alpha L^{-1} , \Sigma_0^{ss}=\alpha L^{-1} $ (
% $\mSigma_0 = \mathrm{diag}(\Sigma_0^{qq}, \Sigma_0^{pp}, \Sigma_0^{ss}) \otimes \mI_d$
% ), where $\alpha \ll  1$. Then $p(\rvx_t \mid \rvx_0, t) = p(\rvx_t \mid \rvq_0, t)$.

\subsection{Lie-Trotter HOLD Sampler}

After we obtain the optimal predictor 
$\mathfrak{S}_\vtheta(\rvx_t, t)$ 
for score $\nabla_{\rvs_t} \log p(\rvx_t,t)$, 
which can be plugged into the backward HOLD for denoising the prior 
distribution into the target data.
The backward or the generative HOLD is approximated as 
\begin{align}
 d\rvq_t & = -\rvp_t dt, \\
 d\rvp_t & =  \rvq_t  dt  -  \gamma\rvs_t dt, \\
 d\rvs_t & =  \gamma\rvp_tdt + \xi\rvs_t dt 
 + 2\xi L^{-1} \mathfrak{S}_\vtheta(\rvx_t, t)dt
 + \sqrt{2\xi L^{-1}} d\bar\rvw_t.
\end{align}
or the matrix form
\begin{equation} 
  \begin{pmatrix} d\rvq_t \\ d\rvp_t \\ d\rvs_t \end{pmatrix} 
  =\begin{pmatrix}  \bm{0}_d \\ \rvq_t \\\bm{0}_d \end{pmatrix} dt 
  +  \begin{pmatrix}  -\rvp_t \\ \bm{0}_d \\ \gamma \rvp_t \end{pmatrix} dt +
  \begin{pmatrix}\bm{0}_d \\ -\gamma \rvs_t \\  \xi\rvs_t\end{pmatrix} dt +
  \begin{pmatrix}\bm{0}_d \\ \bm{0}_d \\ 2\xi L^{-1}\mathfrak{S}_\vtheta(\rvx_t, t) \end{pmatrix}dt +
  \begin{pmatrix}\bm{0}_d \\ \bm{0}_d \\ \sqrt{2\xi L^{-1}} d\bar \rvw_t \end{pmatrix}
\end{equation}
or
\begin{equation} 
  \begin{pmatrix} d\rvq_t \\ d\rvp_t \\ d\rvs_t \end{pmatrix} 
  =\begin{pmatrix}  \bm{0}_d \\ \rvq_t \\\bm{0}_d \end{pmatrix} dt 
  +  \begin{pmatrix}  -\rvp_t \\ \bm{0}_d \\ \gamma \rvp_t \end{pmatrix} dt +
  \begin{pmatrix}\bm{0}_d \\ -\gamma \rvs_t \\  -\xi\rvs_t\end{pmatrix} dt +
  \begin{pmatrix}\bm{0}_d \\ \bm{0}_d \\ \sqrt{2\xi L^{-1}} d\bar \rvw_t \end{pmatrix}+
  \begin{pmatrix}\bm{0}_d \\ \bm{0}_d \\ 2\xi \left[L^{-1}\mathfrak{S}_\vtheta(\rvx_t, t) +\rvs_t \right] \end{pmatrix}dt.
\end{equation}
In this paper, the sampling problem is solved by the 
Lie-Trotter method, also known as the split operator method. 
The basic idea in a nutshell, that is, the complex operator 
is divided into several parts, and then the corresponding 
flow map of each part is determined, and then these maps 
are composed to obtain the numerical algorithm of the original 
complex operator.
It can be seen that this method is particularly suitable for 
complex scenarios without exact closed-form solution

We divide the time reverse HOLD into two parts,
where the first part (A) can be sampled accurately, 
and the second part (B)
can be obtained with high precision approximate solution 
by numerical algorithm.
Of course there are two arrangements, ABA or BAB. 
We choose to put the part of solving B in the middle, 
so that there is no need to solve it multiple times, 
and there is less error than putting it on both sides.
After taking n iterations, we get the form of ABABABABA$\cdots$.

The evolution of the probability distribution 
$p(\rvx_t, t)$ for this backward HOLD is 
described by the general Fokker--Planck 
equation~\cite{sarkka2019applied}:
\begin{align}
  \frac{\partial p(\rvx_t,t)}{\partial t}  = 
  -\sum_{i=1}^{3d}\frac{\partial \left[ (\rva(\rvx_t,t))_i p(\rvx_t,t)\right]}{\partial x_i}  
  +\sum_{i=1}^{3d}\sum_{j=1}^{3d} \frac{\partial^2 }{\partial x_i \partial x_j } 
  [(\mB )_{i,j}p(\rvx_t,t)],
  \label{eq:backward_fpk_pde}
 \end{align}
with
\begin{align}
  \begin{split} 
  \rva(\rvx_t, t) & = 
  \begin{pmatrix}  \bm{0}_d \\ \rvq_t \\\bm{0}_d \end{pmatrix}  
  +  \begin{pmatrix}  -\rvp_t \\ \bm{0}_d \\ \gamma \rvp_t \end{pmatrix}  +
  \begin{pmatrix}\bm{0}_d \\ -\gamma \rvs_t \\  -\xi\rvs_t\end{pmatrix}  +
  \begin{pmatrix}\bm{0}_d \\ \bm{0}_d \\ 2\xi \left[L^{-1}\mathfrak{S}_\vtheta(\rvx_t, t) +\rvs_t \right] \end{pmatrix},\\
  \mB & =
  \begin{pmatrix} 0 &0 &0\\ 
    0 &0 &0\\ 
    0  & 0 &\xi L^{-1} \end{pmatrix} \otimes \mI_d.
\end{split}
\end{align} 
Expanding and simplifying the above equation, we obtain 
\begin{align}
  & \frac{\partial p(\rvx_t,t)}{\partial t} \\ = &
  -\sum_{i=1}^{d}\left[\frac{\partial \left[ (-\rvp_t)_i p(\rvx_t,t)\right]}{\partial q_i} 
  + \frac{\partial \left[ (\rvq_t-\gamma \rvs_t)_i p(\rvx_t,t)\right]}{\partial p_i} 
   + \frac{\partial \left[ (\gamma \rvp_t -\xi\rvs_t + 2\xi \left[L^{-1}\mathfrak{S}_\vtheta(\rvx_t, t) +\rvs_t \right])_i p(\rvx_t,t)\right]}{\partial s_i}  \right]
  \\ 
  +&\sum_{i=1}^{d} \xi L^{-1} \frac{\partial^2 }{\partial s_i^2} p(\rvx_t,t).
  \label{eq:backward_fpk_pde_expanded}
 \end{align}
So we can write the Fokker--Planck 
equation in short form as
\begin{align}
  \frac{\partial p(\rvx_t,t)}{\partial t} 
 = (\mathcal{L}_{A}^\dagger{+}\mathcal{L}_{B}^\dagger)p(\rvx_t,t),
 \label{eq:backward_fpk_pde_short}
\end{align}
where
\begin{align}
  \begin{split} 
    \mathcal{L}_{A}^\dagger(p(\rvx_t,t)) & = -\sum_{i=1}^{d}\left[ (-\rvp_t)_i \frac{\partial  p(\rvx_t,t)}{\partial q_i} 
    + (\rvq_t-\gamma \rvs_t)_i \frac{\partial  p(\rvx_t,t)}{\partial p_i} 
     + (\gamma \rvp_t -\xi\rvs_t)_i\frac{\partial   p(\rvx_t,t)}{\partial s_i}  \right]
    +\sum_{i=1}^{d} \xi L^{-1} \frac{\partial^2 }{\partial s_i^2} p(\rvx_t,t),\\
    \mathcal{L}_{B}^\dagger(p(\rvx_t,t)) & = -2\xi \nabla_{\rvs_t}\left[\left(L^{-1}\mathfrak{S}_\vtheta(\rvx_t, t) +\rvs_t\right) p(\rvx_t,t) \right].
\end{split}
\end{align}
Since $\mathcal{L}_{A}^\dagger$ and $\mathcal{L}_{B}^\dagger$ 
are linear operators, 
thus the solution of Eq.~(\ref{eq:backward_fpk_pde_short}) is 
$p(\rvx_t,t) = e^{(\mathcal{L}_{A}^\dagger+\mathcal{L}_{B}^\dagger)t}p(\rvx_0,0)$~\cite{strang1968construction}.
% \[
%   \rvx_t = e^{(\mathcal{L}_{A}^\dagger+\mathcal{L}_{B}^\dagger)t}\rvx_0
% \]
If $\mathcal{L}_{A}^\dagger$ and $\mathcal{L}_{B}^\dagger$ commute, then by the 
exponential laws we have the equivalent form $p(\rvx_t,t) = e^{t\mathcal{L}_{A}^\dagger}e^{t\mathcal{L}_{B}^\dagger}p(\rvx_0,0)$.
% \[
%   \rvx_t = e^{\mathcal{L}_{A}^\daggert}e^{\mathcal{L}_{B}^\daggert}\rvx_0
% \]
If they do not commute, then by the Baker-Campbell-Hausdorff 
formula~\cite{campbell1896law,hausdorff1906symbolische,baker1901further} it 
is still possible to replace the exponential of the sum by a 
product of exponentials at the cost of a first-order error:
\[
  e^{(\mathcal{L}_{A}^\dagger+\mathcal{L}_{B}^\dagger)t}\rvx_0
  {=}e^{t\mathcal{L}_{A}^\dagger}e^{t\mathcal{L}_{B}^\dagger}\rvx_0+O(t).
\]
This gives rise to a numerical scheme where one, instead of solving 
the original initial problem, solves both subproblems alternating:
\begin{equation}
  \begin{split}
    \tilde{p}(\rvx_{\Delta t},\Delta t) = e^{\mathcal{L}_{A}^\dagger\Delta t}p(\rvx_0,0) \\
    p(\rvx_{\Delta t},\Delta t) = e^{\mathcal{L}_{B}^\dagger\Delta t}\tilde{p}(\rvx_{\Delta t},\Delta t)
  \end{split}
\end{equation}
Strang splitting~\cite{strang1968construction} extends this 
approach 
to second order by choosing 
another order of operations. Instead of taking full time steps 
with each operator, instead, one performs time steps as follows:
\begin{equation}
  \begin{split}
    \tilde{p}(\rvx_{\Delta t},\Delta t) = e^{\mathcal{L}_{A}^\dagger\frac{\Delta t}{2}}p(\rvx_0,0), \\
    \bar{p}(\rvx_{\Delta t},\Delta t) = e^{\mathcal{L}_{B}^\dagger\Delta t} \tilde{p}(\rvx_{\Delta t},\Delta t),  \\
  p(\rvx_{\Delta t},\Delta t) = e^{\mathcal{L}_{A}^\dagger\frac{\Delta t}{2}}\bar{p}(\rvx_{\Delta t},\Delta t).
  \end{split}
\end{equation}
One can prove that Strang splitting is of second order by using 
either the Baker-Campbell-Hausdorff formula, rooted tree 
analysis or a direct comparison of the error terms using Taylor 
expansion.  This paper adopts this algorithm for HOLD-DGM's sampler.

Then we introduce how these two operators $e^{\mathcal{L}_{A}^\dagger\frac{\Delta t}{2}}$ and $e^{\mathcal{L}_{B}^\dagger\Delta t}$ are calculated.

\subsection{Calculation of $e^{\mathcal{L}_{A}^\dagger\frac{\Delta t}{2}}$}
\label{sec:app_calc_op_a}

$e^{\mathcal{L}_{A}^\dagger\frac{\Delta t}{2}}$ describes the evolution under
\begin{equation} 
  \begin{pmatrix} d\rvq_t \\ d\rvp_t \\ d\rvs_t \end{pmatrix} 
  =\begin{pmatrix}  \bm{0}_d \\ \rvq_t \\\bm{0}_d \end{pmatrix} dt 
  +  \begin{pmatrix}  -\rvp_t \\ \bm{0}_d \\ \gamma \rvp_t \end{pmatrix} dt +
  \begin{pmatrix}\bm{0}_d \\ -\gamma \rvs_t \\  -\xi\rvs_t\end{pmatrix} dt +
  \begin{pmatrix}\bm{0}_d \\ \bm{0}_d \\ \sqrt{2\xi L^{-1}} d\bar \rvw_t \end{pmatrix}.
\label{eq:app_sde_a}
\end{equation}
This problem is very similar (but different) to the previous 
problem~(\ref{eq:app_hold_matrix}), 
thus we quickly go through it.
Let 
\begin{align}
  \vl(t) \coloneqq  \begin{pmatrix} 0 &-1&0\\ 
    1 &0&-\sqrt{10}\\ 
    0 &  \sqrt{10} & -6 \end{pmatrix} \otimes \mI_d,
    \qquad
  \mM(t) \coloneqq \begin{pmatrix} 0 &0 &0\\ 
    0 &0 &0\\ 
    0  & 0 &\sqrt{12 L^{-1}} \end{pmatrix} \otimes \mI_d.
\end{align}
The transition densities 
$p(\rvx_t|\rvx_0)$ of the solution 
process $\rvx_t$ for the SDE~(\ref{eq:app_sde_a}) 
is the solution to the Fokker-Planck-Kolmogorov (FPK)
equation~\cite{sarkka2019applied}
\begin{align}
  \frac{\partial p(\rvx_t,t)}{\partial t}  = 
  -\sum_{i=1}^d\frac{\partial \left[ (\vl(t)\rvx_t)_i p(\rvx_t,t)\right]}{\partial x_i}  
  +\sum_{i=1}^d\sum_{j=1}^d \frac{\partial^2 }{\partial x_i \partial x_j } 
  [(\mM(t) \mM(t)^\top)_{i,j}p(\rvx_t,t)].
  \label{eq:fpk_pde_sampling}
 \end{align}
$p(\mathbf{x}(t)|\mathbf{x}(0))$ 
is Gaussian with mean $\vmu_t$ and variance
$\mSigma_t$ satisfy the following ODEs~\cite{sarkka2019applied}
\begin{align} 
 \frac{d\vmu_t}{dt} &= \vl(t) \vmu_t, \label{app:ode_mean_sampling} \\
 \frac{d\mSigma_t}{dt} &=\vl(t) \mSigma_t + \left[ \vl(t) \mSigma_t \right]^\top + 
 \mM(t) \mM(t)^\top . \label{app:ode_var_sampling}
\end{align}
This array of ODE is easy to solve, and we have
\begin{align} 
  \vmu_{t+\frac{\Delta t}{2}} &
  = \exp\left[\int_t^{t+\frac{\Delta t}{2}}\vl(\tau)d\tau \right] \vmu_{t}, \\
  \mSigma_{t+\frac{\Delta t}{2}} &= \exp\left[ \int_t^{t+\frac{\Delta t}{2}}\vl(\tau)d\tau \right] \mSigma_t 
  \exp\left[ \int_t^{t+\frac{\Delta t}{2}}\vl(\tau)d\tau \right]^\top  
   \\& +  \int_t^{t+\frac{\Delta t}{2}}  \exp\left[ \int_s^{t+\frac{\Delta t}{2}} \vl(\tau)d\tau \right] \mM(s) \mM(s)^\top 
  \exp\left[ \int_s^{t+\frac{\Delta t}{2}} \vl(\tau)d\tau  \right]^\top ds
  \label{eq:app_mean_var_sampling}
\end{align}
where
\begin{align} 
  \int_t^{t+\frac{\Delta t}{2}}\vl(\tau)d\tau =  
  \frac{\Delta t}{2} \begin{pmatrix} 0 &-1&0\\ 
    1 &0&-\sqrt{10}\\ 
    0 & \sqrt{10} & -6 \end{pmatrix} \otimes \mI_d.
\end{align}
  Let $\mL =  \begin{pmatrix} 0 &-1&0\\ 
     1 &0&-\sqrt{10}\\ 
     0 & \sqrt{10} & -6 \end{pmatrix}  $, then the eigenvalues 
     of  $\mL$ are  $-3$,  $-2$, and
        $-1$. By the Putzer’s spectral 
  formula~\cite{putzer1966avoiding} and Lemma~\ref{lem:ekp} 
  we have
  \begin{align}
    \label{eq:L_put_sampling}
    \exp\left[\int_t^{t+\frac{\Delta t}{2}}\vl(\tau)d\tau \right]
    =\exp(\frac{\Delta t}{2}\mL\otimes \mI_d)
    =\left[r_1(\frac{\Delta t}{2})\mP_1+r_2(\frac{\Delta t}{2})\mP_2+r_3(\frac{\Delta t}{2})\mP_3\right]\otimes \mI_d,
  \end{align}
  where
  \begin{align}
     \mP_1=\mI, \quad \mP_2=(\mL+3\mI)\mP_1, \quad \mP_3=(\mL+2\mI)\mP_2, 
   \end{align}
     and 
   \begin{align}
     &r'_1(\frac{\Delta t}{2})=-3r_1(\frac{\Delta t}{2}), \quad r_1(0)=1, \\
     &r'_2(\frac{\Delta t}{2})=-2r_2(\frac{\Delta t}{2})+r_1(t), \quad r_2(0)=0,\\
     &r'_3(\frac{\Delta t}{2})=-r_3(\frac{\Delta t}{2})+r_2(\frac{\Delta t}{2}), \quad r_3(0)=0. 
   \end{align}
Therefore
\begin{equation} 
 \begin{split}
 &\mP_1=\mI, \quad
 \mP_2=\begin{pmatrix} 3 &-1&0\\ 
   1 &3&-\sqrt{10}\\ 
   0 & \sqrt{10} & -3 \end{pmatrix},  \quad
   \mP_3=\begin{pmatrix} 5&-5&\sqrt{10}\\ 
   5 & 5 & \sqrt{10}\\ 
   \sqrt{10} & -\sqrt{10} & 2 \end{pmatrix},   \label{eq:mP_sampling}
 \end{split}
\end{equation} 
and
\begin{equation} 
  \begin{split}
  &r_1(\frac{\Delta t}{2})=\exp\left(-3\frac{\Delta t}{2}t\right),   \quad 
  r_2(\frac{\Delta t}{2})=\exp\left(-2\frac{\Delta t}{2}\right) - \exp\left(-3\frac{\Delta t}{2}\right),\\
  &r_3(\frac{\Delta t}{2})=\frac{1}{2}\exp\left(-\frac{\Delta t}{2}\right)  
  + \frac{1}{2}\exp\left(-3\frac{\Delta t}{2}\right) - \exp\left(-2\frac{\Delta t}{2}\right).  \label{eq:r_sampling}
  \end{split}
\end{equation} 
Plugging~(\ref{eq:mP_sampling}) and~(\ref{eq:r_sampling}) 
into~(\ref{eq:L_put_sampling}), 
we obtain the elements of $\exp(\frac{\Delta t}{2}\mL)$ are
\begin{align}
&l_{11}(\frac{\Delta t}{2})= \frac{5}{2}\exp(-\frac{\Delta t}{2}) -2\exp(-2\frac{\Delta t}{2}) + \frac{1}{2}\exp(-3\frac{\Delta t}{2}), \\
&l_{12}(\frac{\Delta t}{2})= -\frac{5}{2}\exp(-\frac{\Delta t}{2}) + 4\exp(-2\frac{\Delta t}{2}) - \frac{3}{2}\exp(-3\frac{\Delta t}{2}), \\
&l_{13}(\frac{\Delta t}{2})= \sqrt{10}\left[\frac{1}{2}\exp\left(-\frac{\Delta t}{2}\right)  - \exp\left(-2t\right) + \frac{1}{2}\exp\left(-3\frac{\Delta t}{2}\right) \right], \\
&l_{21}(\frac{\Delta t}{2})=  \frac{5}{2}\exp(-\frac{\Delta t}{2}) - 4\exp(-2\frac{\Delta t}{2}) + \frac{3}{2}\exp(-3\frac{\Delta t}{2}), \\
&l_{22}(\frac{\Delta t}{2})= \frac{5}{2}\exp(-\frac{\Delta t}{2}) -2\exp(-2\frac{\Delta t}{2}) + \frac{1}{2}\exp(-3\frac{\Delta t}{2}), \\ 
&l_{23}(\frac{\Delta t}{2})= \sqrt{10}\left[ \frac{1}{2}\exp\left(-\frac{\Delta t}{2}\right) - 2\exp\left(-2\frac{\Delta t}{2}\right) + \frac{3}{2}\exp\left(-3\frac{\Delta t}{2}\right) \right], \\
&l_{31}(\frac{\Delta t}{2})= \sqrt{10}\left[\frac{1}{2}\exp\left(-\frac{\Delta t}{2}\right)  - \exp\left(-2\frac{\Delta t}{2}\right) + \frac{1}{2}\exp\left(-3\frac{\Delta t}{2}\right) \right], \\
&l_{32}(\frac{\Delta t}{2})= \sqrt{10}\left[ -\frac{1}{2}\exp\left(-\frac{\Delta t}{2}\right)  + 2\exp\left(-2\frac{\Delta t}{2}\right) - \frac{3}{2}\exp\left(-3\frac{\Delta t}{2}\right) \right], \\
&l_{33}(\frac{\Delta t}{2})= \exp(-\frac{\Delta t}{2}) -5\exp(-2\frac{\Delta t}{2}) + 5\exp(-3\frac{\Delta }{2}). 
\end{align}
Let  $\vmu_{t} = (\rvq_{t}, \rvp_{t}, \rvs_{t})^\top$, thus the 
mean $\vmu_t$ is 
\begin{align}
 \vmu_{t+\frac{\Delta t}{2}} = 
 \begin{pmatrix} l_{11}(\frac{\Delta t}{2})\rvq_{t} + l_{12}(\frac{\Delta t}{2})\rvp_{t} + l_{13}(\frac{\Delta t}{2})\rvs_{t} \\ 
   l_{21}(\frac{\Delta t}{2})\rvq_{t} + l_{22}(\frac{\Delta t}{2})\rvp_{t} + l_{23}(\frac{\Delta t}{2})\rvs_{t} \\
   l_{31}(\frac{\Delta t}{2})\rvq_{t} + l_{32}(\frac{\Delta t}{2})\rvp_{t} + l_{33}(\frac{\Delta t}{2})\rvs_{t} \end{pmatrix}.  
   \label{eq:mean_sampling}
\end{align}
For the variance, let  
$\mSigma_t = \mathrm{diag}(\Sigma_t^{qq}, \Sigma_t^{pp}, \Sigma_t^{ss}) \otimes \mI_d$, plugging the 
formula of $\exp(t\mL)$ (Eq.~(\ref{eq:L_put_sampling})) into 
$\mSigma_{t+\frac{\Delta t}{2}}$ (Eq.~(\ref{eq:app_mean_var_sampling})), 
implies that 
$\mSigma_{t+\frac{\Delta t}{2}}$ equals
\begin{align} 
& \exp\left[ \int_t^{t+\frac{\Delta t}{2}}\vl(\tau)d\tau \right] \mSigma_t 
 \exp\left[ \int_t^{t+\frac{\Delta t}{2}}\vl(\tau)d\tau \right]^\top   
 +  \int_t^{t+\frac{\Delta t}{2}}  \exp\left[ \int_s^{t+\frac{\Delta t}{2}} \vl(\tau)d\tau \right] \mM(s) \mM(s)^\top 
 \exp\left[ \int_s^{t+\frac{\Delta t}{2}}\vf(\tau)d\tau  \right]^\top ds \\
 &= \left[\begin{pmatrix} l_{11}(\frac{\Delta t}{2}) &l_{12}(\frac{\Delta t}{2})&l_{13}(\frac{\Delta t}{2})\\ 
   l_{21}(\frac{\Delta t}{2}) &l_{22}(\frac{\Delta t}{2})&l_{23}(\frac{\Delta t}{2})\\ 
   l_{31}(\frac{\Delta t}{2}) &l_{32}(\frac{\Delta t}{2})&l_{33}(\frac{\Delta t}{2})  \end{pmatrix}  
   \begin{pmatrix} \Sigma_t^{qq} &0&0\\ 
     0 &\Sigma_t^{pp}&0\\ 
     0 &0&\Sigma_t^{ss}  \end{pmatrix}  
     \begin{pmatrix} l_{11}(\frac{\Delta t}{2}) &l_{12}(\frac{\Delta t}{2})&l_{13}(\frac{\Delta t}{2})\\ 
      l_{21}(\frac{\Delta t}{2}) &l_{22}(\frac{\Delta t}{2})&l_{23}(\frac{\Delta t}{2})\\ 
      l_{31}(\frac{\Delta t}{2}) &l_{32}(\frac{\Delta t}{2})&l_{33}(\frac{\Delta t}{2})  \end{pmatrix}^\top\right] \otimes \mI_d \\
&+\int_{t}^{t+\frac{\Delta t}{2}} \left[ 
  \begin{pmatrix} l_{11}(t\frac{\Delta t}{2}s) &l_{12}(t\frac{\Delta t}{2}s)&l_{13}(t\frac{\Delta t}{2}s)\\ 
 l_{21}(t\frac{\Delta t}{2}s) &l_{22}(t\frac{\Delta t}{2}s)&l_{23}(t\frac{\Delta t}{2}s)\\ 
 l_{31}(t\frac{\Delta t}{2}s) &l_{32}(t\frac{\Delta t}{2}s)&l_{33}(t\frac{\Delta t}{2}s)  \end{pmatrix} 
 \begin{pmatrix} 0 &0&0\\ 
 0 &0&0\\ 
0 &0&12 L^{-1}  \end{pmatrix} 
\begin{pmatrix} l_{11}(t\frac{\Delta t}{2}s) &l_{12}(t\frac{\Delta t}{2}s)&l_{13}(t\frac{\Delta t}{2}s)\\ 
  l_{21}(t\frac{\Delta t}{2}s) &l_{22}(t\frac{\Delta t}{2}s)&l_{23}(t\frac{\Delta t}{2}s)\\ 
  l_{31}(t\frac{\Delta t}{2}s) &l_{32}(t\frac{\Delta t}{2}s)&l_{33}(t\frac{\Delta t}{2}s)  \end{pmatrix} ^\top\right] \otimes \mI_d ds \\
 &= \begin{pmatrix} 
  \sum_{j=1}^3 l_{1j}^2(\frac{\Delta t}{2})\Sigma_t^{jj} & \sum_{j=1}^3 l_{1j}(\frac{\Delta t}{2})l_{2j}(\frac{\Delta t}{2})\Sigma_t^{jj} & \sum_{j=1}^3 l_{1j}(\frac{\Delta t}{2})l_{3j}(\frac{\Delta t}{2})\Sigma_t^{jj} \\ 
   \sum_{j=1}^3 l_{2j}(\frac{\Delta t}{2})l_{1j}(\frac{\Delta t}{2})\Sigma_t^{jj} & \sum_{j=1}^3 l_{2j}^2(\frac{\Delta t}{2})\Sigma_t^{jj} & \sum_{j=1}^3 l_{2j}(\frac{\Delta t}{2})l_{3j}(\frac{\Delta t}{2})\Sigma_t^{jj} \\ 
   \sum_{j=1}^3 l_{3j}(\frac{\Delta t}{2})l_{1j}(\frac{\Delta t}{2})\Sigma_t^{jj} & \sum_{j=1}^3 l_{3j}(\frac{\Delta t}{2})l_{2j}(\frac{\Delta t}{2})\Sigma_t^{jj} & \sum_{j=1}^3 l_{3j}^2(\frac{\Delta t}{2})\Sigma_t^{jj}  \end{pmatrix}   \otimes \mI_d \\
&+ 12 L^{-1} \int_{t}^{t+\frac{\Delta t}{2}}  
\begin{pmatrix}  l_{13}^2(t\frac{\Delta t}{2}s) &  l_{13}(t\frac{\Delta t}{2}s)l_{23}(t\frac{\Delta t}{2}s) &  l_{13}(t\frac{\Delta t}{2}s)l_{33}(t\frac{\Delta t}{2}s) \\ 
  l_{23}(t\frac{\Delta t}{2}s)l_{13}(t\frac{\Delta t}{2}s) &  l_{23}^2(t\frac{\Delta t}{2}s) &  l_{23}(t\frac{\Delta t}{2}s)l_{33}(t\frac{\Delta t}{2}s) \\ 
l_{33}(t\frac{\Delta t}{2}s)l_{13}(t\frac{\Delta t}{2}s) & l_{33}(t\frac{\Delta t}{2}s)f_{23}(t\frac{\Delta t}{2}s) &  l_{33}^2(t\frac{\Delta t}{2}s)  \end{pmatrix}  ds  \otimes \mI_d
%&+ 12 L^{-1} \int_{t}^{t+\frac{\Delta t}{2}}  
% \begin{pmatrix}  l_{13}^2(t+\frac{\Delta t}{2}-s) &  l_{13}(t+\frac{\Delta t}{2}-s)l_{23}(t+\frac{\Delta t}{2}-s) &  l_{13}(t+\frac{\Delta t}{2}-s)l_{33}(t+\frac{\Delta t}{2}-s) \\ 
%   l_{23}(t+\frac{\Delta t}{2}-s)l_{13}(t+\frac{\Delta t}{2}-s) &  l_{23}^2(t+\frac{\Delta t}{2}-s) &  l_{23}(t+\frac{\Delta t}{2}-s)l_{33}(t+\frac{\Delta t}{2}-s) \\ 
% l_{33}(t+\frac{\Delta t}{2}-s)l_{13}(t+\frac{\Delta t}{2}-s) & l_{33}(t+\frac{\Delta t}{2}-s)l_{23}(t+\frac{\Delta t}{2}-s) &  l_{33}^2(t+\frac{\Delta t}{2}-s)  \end{pmatrix}  ds  \otimes \mI_d
\end{align}
where we let $1\leftrightarrow q, 2\leftrightarrow p, 3\leftrightarrow s$ in $\Sigma_0^{jj}$ to simplify notation
, and $t\frac{\Delta t}{2}s=t+\frac{\Delta t}{2}-s$.
On the other hand, we have 
\begin{align}
\mSigma_{t+\frac{\Delta t}{2}} = \Sigma_{t+\frac{\Delta t}{2}}  \otimes \mI_d, \quad \Sigma_{t+\frac{\Delta t}{2}}  = 
\begin{pmatrix} \Sigma_{t+\frac{\Delta t}{2}}^{qq} & \Sigma_{t+\frac{\Delta t}{2}}^{qp} & \Sigma_{t+\frac{\Delta t}{2}}^{qs}\\ 
 \Sigma_{t+\frac{\Delta t}{2}}^{qp} & \Sigma_{t+\frac{\Delta t}{2}}^{pp} & \Sigma_{t+\frac{\Delta t}{2}}^{ps}\\ 
 \Sigma_{t+\frac{\Delta t}{2}}^{qs} & \Sigma_{t+\frac{\Delta t}{2}}^{ps} & \Sigma_{t+\frac{\Delta t}{2}}^{ss}  \end{pmatrix},
\end{align}
which clearly implies that
\begin{align}
 &\Sigma_{t+\frac{\Delta t}{2}}^{qq} = \sum_{j=1}^3 l_{1j}^2(\frac{\Delta t}{2})\Sigma_t^{jj} +12 L^{-1} \int_{t}^{t+\frac{\Delta t}{2}}  l_{13}^2(t+\frac{\Delta t}{2}-s) ds , \\
 &\Sigma_{t+\frac{\Delta t}{2}}^{qp}=  \sum_{j=1}^3 l_{1j}(\frac{\Delta t}{2})l_{2j}(\frac{\Delta t}{2})\Sigma_t^{jj} +  12 L^{-1} \int_{t}^{t+\frac{\Delta t}{2}}  l_{13}(t+\frac{\Delta t}{2}-s)l_{23}(t+\frac{\Delta t}{2}-s) ds, \\
 &\Sigma_{t+\frac{\Delta t}{2}}^{qs}= \sum_{j=1}^3 l_{1j}(\frac{\Delta t}{2})l_{3j}(\frac{\Delta t}{2})\Sigma_t^{jj} + 12 L^{-1}  \int_{t}^{t+\frac{\Delta t}{2}}  l_{13}(t+\frac{\Delta t}{2}-s)l_{33}(t+\frac{\Delta t}{2}-s)  ds, \\
 &\Sigma_{t+\frac{\Delta t}{2}}^{pp}= \sum_{j=1}^3 l_{2j}^2(\frac{\Delta t}{2})\Sigma_t^{jj} + 12 L^{-1}  \int_{t}^{t+\frac{\Delta t}{2}}  l_{23}^2(t+\frac{\Delta t}{2}-s) ds, \\
 &\Sigma_{t+\frac{\Delta t}{2}}^{ps}= \sum_{j=1}^3 l_{2j}(\frac{\Delta t}{2})l_{3j}(\frac{\Delta t}{2})\Sigma_t^{jj}  + 12 L^{-1}  \int_{t}^{t+\frac{\Delta t}{2}}   l_{23}(t+\frac{\Delta t}{2}-s)l_{33}(t+\frac{\Delta t}{2}-s) ds, \\ 
 &\Sigma_{t+\frac{\Delta t}{2}}^{ss}= \sum_{j=1}^3 l_{3j}^2(\frac{\Delta t}{2})\Sigma_t^{jj}  + 12 L^{-1}  \int_{t}^{t+\frac{\Delta t}{2}}   l_{33}^2(t+\frac{\Delta t}{2}-s)ds,
 \end{align}
where
\begin{align}
  &l_{13}^2(t)= l_{31}^2(t)=\frac{5}{2}\exp(-2t) - 10\exp(-3t) + 15\exp(-4t)  - 10\exp(-5t)+\frac{5}{2}\exp(-6t), \\
  &l_{23}^2(t)= l_{32}^2(t)= \frac{5}{2}\exp(-2t) - 20\exp(-3t) + 55\exp(-4t)  - 60\exp(-5t)+\frac{45}{2}\exp(-6t), \\
  &l_{33}^2(t)=  \exp(-2t) - 10 \exp(-3t) + 35\exp(-4t)  - 50\exp(-5t)+25\exp(-6t), \\
 % \end{align}
 % \begin{align}
  &l_{13}(t)l_{23}(t)=\frac{5}{2}\exp(-2t) -15\exp(-3t) +30\exp(-4t)  - 25\exp(-5t)+\frac{15}{2}\exp(-6t), \\
  &l_{13}(t)l_{33}(t)=\frac{\sqrt{10}}{2}\exp(-2t) -\frac{7\sqrt{10}}{2}\exp(-3t) + 8\sqrt{10}\exp(-4t)  - \frac{15\sqrt{10}}{2}\exp(-5t)+\frac{5\sqrt{10}}{2}\exp(-6t), \\
  &l_{23}(t)l_{33}(t)= \frac{\sqrt{10}}{2}\exp(-2t) -\frac{9\sqrt{10}}{2}\exp(-3t) + 14\sqrt{10}\exp(-4t)  
  - \frac{35\sqrt{10}}{2}\exp(-5t)+\frac{15\sqrt{10}}{2}\exp(-6t), \\
%  \end{align}
%    %
%    \begin{align}
    &l_{11}^2(t)= \frac{25}{4}\exp(-2t) - 10\exp(-3t) + \frac{13}{2}\exp(-4t) - 2\exp(-5t)+\frac{1}{4}\exp(-6t), \\
    &l_{12}^2(t)= l_{21}^2(t)=\frac{25}{4}\exp(-2t) - 20\exp(-3t) + \frac{47}{2}\exp(-4t) - 12\exp(-5t)+\frac{9}{4}\exp(-6t),\\
    &l_{22}^2(t)= \frac{25}{4}\exp(-2t) - 10\exp(-3t) + \frac{13}{2}\exp(-4t)  - 2\exp(-5t)+\frac{1}{4}\exp(-6t), \\
  \end{align}
  \begin{align}
    &l_{21}(t)l_{31}(t)=\frac{5\sqrt{10}}{4}\exp(-2t) -\frac{9\sqrt{10}}{2}\exp(-3t) + 6\sqrt{10}\exp(-4t)  
    - \frac{7\sqrt{10}}{2}\exp(-5t)+\frac{3\sqrt{10}}{4}\exp(-6t), \\
    &l_{22}(t)l_{32}(t)=\sqrt{10}\left[-\frac{5}{4}\exp(-2t) 
    + 6\exp(-3t) -8\exp(-4t)  
    + 4\exp(-5t)-\frac{3}{4}\exp(-6t)\right], \\
    &l_{11}(t)l_{31}(t)= \frac{5\sqrt{10}}{4}\exp(-2t) -\frac{7\sqrt{10}}{2}\exp(-3t) + \frac{7\sqrt{10}}{2}\exp(-4t)  - \frac{3\sqrt{10}}{2}\exp(-5t)+\frac{\sqrt{10}}{4}\exp(-6t), \\ 
   % \end{align}
   % \begin{align}
    &l_{12}(t)l_{32}(t)=\frac{5\sqrt{10}}{4}\exp(-2t) -7\sqrt{10}\exp(-3t) + \frac{25\sqrt{10}}{2}\exp(-4t)  - 9\sqrt{10}\exp(-5t)+\frac{9\sqrt{10}}{4}\exp(-6t), \\
    &l_{11}(t)l_{21}(t)=\frac{25}{4}\exp(-2t) - 15\exp(-3t) +13\exp(-4t)  -5\exp(-5t)+\frac{3}{4}\exp(-6t), \\
    &l_{12}(t)l_{22}(t)=-\frac{25}{4}\exp(-2t) + 15\exp(-3t) -13\exp(-4t)  +5\exp(-5t)-\frac{3}{4}\exp(-6t).
\end{align}

\subsection{Calculation of $e^{\mathcal{L}_{B}^\dagger\Delta t}$}

$\mathcal{L}_{B}^\dagger$ is represented by
\begin{equation} 
  \begin{pmatrix} d\rvq_t \\ d\rvp_t \\ d\rvs_t \end{pmatrix} 
  =
  \begin{pmatrix}\bm{0}_d \\ \bm{0}_d \\ 2\xi \left[L^{-1}\mathfrak{S}_\vtheta(\rvx_t, t) +\rvs_t \right] \end{pmatrix}dt.
\end{equation}
This is an ODE, which can be solved by one-step Euler method
\begin{equation} 
  e^{\mathcal{L}_{B}^\dagger\Delta t} \begin{pmatrix} \rvq_t \\ \rvp_t \\ \rvs_t \end{pmatrix} 
  = \begin{pmatrix} \rvq_t \\ \rvp_t \\ \rvs_t \end{pmatrix}  +
  \begin{pmatrix}\bm{0}_d \\ \bm{0}_d \\ 2\xi \left[L^{-1}\mathfrak{S}_\vtheta(\rvx_t, t) +\rvs_t \right] \end{pmatrix}\Delta t.
\end{equation}
\subsection{Probability Fow ODE and Likelihood Computation}
\label{sec:app_likelihood_comput}

Usually a linear SDE always has a corresponding ODE 
which has the same intermediate marginal distribution 
solution~\cite{sarkka2019applied,song2020score}.
We can approximate the log-likelihood of the given test data 
under this ODE formulation, which defines a 
Normalizing flow~\cite{song2020score}.
\begin{equation} 
  d\rvx_t=K(\rvx_t,t)dt
\end{equation}
  where 
  \begin{equation} 
    K(\rvx_t,t)
    =  \begin{pmatrix} 0 & 1 & 0\\ 
      -1 & 0  & \gamma\\ 
      0 & -\gamma & -\xi \end{pmatrix}
      \rvx_t 
    + \begin{pmatrix} \bm{0}_d \\ \bm{0}_d \\ - \xi L^{-1} \nabla_{\rvs_t} \log p(\rvx_t, t) \end{pmatrix} 
 \end{equation}
  or 
\begin{equation} 
  d\rvx_t=K_\theta(\rvx_t,t)dt
\end{equation}
  where
\begin{equation} 
  K_\theta(\rvx_t,t)
  =  \begin{pmatrix} 0 & 1 & 0\\ 
    -1 & 0  & \gamma\\ 
    0 & -\gamma & -\xi \end{pmatrix}
    \rvx_t 
  + \begin{pmatrix} \bm{0}_d \\ \bm{0}_d \\ - \xi L^{-1} \mathfrak{S}_\vtheta(\rvx_t, t) \end{pmatrix} 
\end{equation}

 In HOLD-DGM, we can compute a 
lower bound on the log-likelihood with
\begin{equation}
  \begin{split}
      \log p(\rvq_0) & = \log \left( \int p(\rvq_0, \rvp_0, \rvs_0) d\rvp_0 d\rvs_0\right) \\
      & = \log \left( \int p(\rvp_0, \rvs_0) \frac{p(\rvq_0, \rvp_0, \rvs_0)}{p(\rvp_0, \rvs_0)}  d\rvp_0 d\rvs_0 \right) \\
      & \geq \E_{\rvp_0, \rvs_0\sim p(\rvp_0, \rvs_0)} \left[\log p(\rvq_0, \rvp_0, \rvs_0) - \log p(\rvp_0, \rvs_0) \right] \\
      & = \E_{\rvp_0, \rvs_0\sim p(\rvp_0, \rvs_0)} \left[\log p(\rvq_0, \rvp_0, \rvs_0)\right] + H(p(\rvp_0, \rvs_0)) \\
      & = \E_{\rvp_0, \rvs_0\sim p(\rvp_0, \rvs_0)} \left[\log p(\rvq_0, \rvp_0, \rvs_0)\right] + H(p(\rvp_0)) + H(p(\rvs_0))  \label{eq:ll_bound}
  \end{split}
  \end{equation}
  where $H(p(\rvp_0))$ and $H(p(\rvs_0))$ 
  denotes the entropy 
  of $p(\rvp_0)$ and $p(\rvs_0)$, and in the derivation of the 
  last `$=$', the Chain 
  rule ($H(p(\rvp_0, \rvs_0))=H(p(\rvp_0|\rvs_0))+H(p(\rvs_0))$) of 
  joint entropy and the fact that $\rvp_0$ and $\rvs_0$ are 
  independent are used.
  Since $p(\rvp_0)\sim\mathcal{N}(\bm{0}_d,\alpha L^{-1}\mI_d)$ 
  and $p(\rvs_0)\sim\mathcal{N}(\bm{0}_d,\alpha L^{-1}\mI_d)$, 
  it is not hard to obtain 
  $H(p(\rvp_0))=H(p(\rvs_0))=\frac{1}{2}+\log{\sqrt{2\pi\alpha L^{-1}}}$. 
  For $\log p(\rvq_0, \rvp_0, \rvs_0)$,
  by~\cite{chen2018neural} and~\cite{song2020score} % (Chen et al., 2018), (Song el al., score sde)
  we use the following stochastic, but 
  unbiased estimate~\cite{chen2018neural,grathwohl2018ffjord}
  \begin{equation}
    \begin{split}
  \log p(\rvq_0, \rvp_0, \rvs_0)
  &= \log p(\rvq_T, \rvp_T, \rvs_T)+
  \int_0^T \nabla\cdot K_\theta(\rvx_t,t) dt \\
  & = \log p(\rvq_T)  +  \log p(\rvp_T) + \log p(\rvs_T)  
  + \int_0^T \nabla\cdot K_\theta(\rvx_t,t) dt  \\
  & \approx \log p(\rvq_T)  +  \log p(\rvp_T) + \log p(\rvs_T) 
  + \int_0^T \rvepsilon^\top \nabla \mJ_{K_\theta}(\rvx_t,t) \rvepsilon dt 
\end{split}
\end{equation}
where $\rvepsilon \sim \gN(\bm{0}, \mI_{d})$ and $\mJ_{K_\theta}(\rvx_t,t)$ is the Jacobian of $K_\theta(\rvx_t,t)$.
Due to  $p(\rvq_T)\sim\mathcal{N}(\bm{0}_d, L^{-1}\mI_d)$,   $p(\rvq_T)\sim\mathcal{N}(\bm{0}_d,L^{-1}\mI_d)$, 
and $p(\rvs_T)\sim\mathcal{N}(\bm{0}_d, L^{-1}\mI_d)$, 
we have $ \log p(\rvq_T)=\log p(\rvp_T)=\log p(\rvs_T)=-\frac{d}{2}\log{2\pi}-\frac{d}{2}\log{L^{-1}} - \frac{L}{2}\parallel \cdot \parallel$. 
We use Runge-Kutta ODE solver of order 5(4) (Dormand \& Prince, 1980) 
in \texttt{torchdiffeq.odeint}  with \texttt{atol=1e-5} 
and \texttt{rtol=1e-5} in all 
situations
to solve the probability flow ODE.
  We report the negative of Eq.~(\ref{eq:ll_bound}) as our upper 
  bound on the negative log-likelihood (NLL).

% \subsection{Probability FLOW Sampling}

\section{Related Work}

Many effective works in continuous DGM adopt VP and VE SDEs~\cite{song2020score} 
as the underlying forward
diffusion 
process~\cite{song2020score,meng2021sdedit,jolicoeur2021gotta,song2021maximum,popov2021diffusion,shi2022iton,kim2021score}.
These two SDEs 
are very classic, and they respectively represent two 
typical changes in the variance of the solution process 
of SDE over time, one tends to a fixed value, and the other tends 
to infinity.
%  (of course, in practice we will not let the 
% variance tend to infinity, but tend to a relatively large 
% fixed value, such as 100~\cite{song2020score}). 
In theory, it is not possible to see which of the two SDEs is 
better, while some experiments show that VE is better than VP 
in the FID measurement for image generation~\cite{song2020score}.
Many applications use these two SDEs directly or 
with  
minor changes, such as speech 
synthesis~\cite{popov2021diffusion,shi2022iton}, 
image editing~\cite{meng2021sdedit}, etc. There are also some works 
trying 
to improve some defects of DGMs based on these two SDEs, 
such as increasing the speed of sampling~\cite{jolicoeur2021gotta},
problem of score divergence as $t$ goes to zero~\cite{kim2021score}.

Our HOLD-based DGM is different from these prior approaches,
which are just tinkering with DGMs based on the VP and VE SDEs.  
HOLD is a brand new SDE that separates position 
from Brownian motion with additional variables, overcoming the 
problems of slow sampling and unbounded scores while improving 
image quality.

%
%\vspace{0.5cm}
   %\vspace{-0.5cm}
   %

The work most related to ours is CLD-SGM~\cite{dockhorn2021score},
which applies score matching with second-order Langevin dynamics
to DGM. 
Our method HOLD is inspired by CLD-SGM, which is a special case of 
ours. HOLD lifts the original 
$2d$-dimensional space to an $nd$-dimensional
space consisting of $n$ vectors of the form,
and considers an $nd$-dimensional
collection of SDEs in these variables.
Our approach extends CLD-SGM over three aspects: 1) HOLD 
generalizes the theoretical framework of CLD-SGM to make it 
more flexible, and any number of auxiliary variables 
can be added to control the smoothness of position variable 
sampling. It can save a lot of computing 
resources without lowering the synthetic performance. 
2) In order to better utilize the capabilities of high-order 
dynamical and reduce the amount of computation, we generalize 
HSM to become a more flexible BCSM, which can flexibly select 
variables that need to be marginalized, and only denoise specific 
variables of interest.  
3) Generalize the Symmetric Splitting CLD Sampler (SSCS) 
so that it can face and handle the 
combination of multiple Hamiltonians and one OU process.
% 4) Although from a formal point of view, HOLD just adds 
% an acceleration
% variable to CLD,  it can save a lot of computing 
% resources without lowering the synthetic performance. 
4) Finally from the perspective of 
computational 
complexity, it increases by an order of magnitude, 
especially the calculation of transition probability 
is quite complicated. Even for the third-order HOLD, 
we spent quite a long time calculating a 
particular solution of the special HOLD, and the calculation 
of the general third-order HOLD, the fourth-order and 
higher-order HOLD will be our future work.

\section{More Experimental Details}
\label{appsec:exp_details}

\subsection{Generate Data from 1D Gaussian Mixtures}
\label{appsec:gen_1d_gm}

In this section, we give details on how to depict
Figure~\ref{fig:1d_diffusion_comparision} 
and Figure~\ref{fig:hold_1d_diffusion_qps} in the main body, 
that is, how to achieve the mutual conversion between a 
single Gaussian distribution and a Gaussian mixture distribution 
based on HOLD in the 1D case. The target distribution is
\begin{align}
  p_\mathrm{data}(\rvx) 
  = 0.34 \gN(\rvx; -0.6575, 0.01^2 \mI_2) 
  +0.33 \gN(\rvx; 0.2474, 0.02^2 \mI_2) 
  + 0.33 \gN(\rvx; 0.8002, 0.01^2 \mI_2). 
\end{align}
The principle of choosing the mean and standard deviation here 
is to make the individual Gaussian components separated far 
enough apart to have different shapes so that they can be 
easily distinguished.
This is a relatively easy generation task, and in the VP, 
CLD, and HOLD experiments, we all use a simple model to 
predict scores.
The model is a 5-layer, fully 
connected network (also called multilayer perceptron, MLP) with 128 neurons in the hidden layers, 
and uses \texttt{nn.SiLU} as the activation function.

In order to draw the background in 
Figure~\ref{fig:1d_diffusion_comparision} 
and Figure~\ref{fig:hold_1d_diffusion_qps}, for 
reverse VP, CLD, HOLD, \texttt{torchdiffeq.odeint} is used to 
calculate 40960 solution 
paths respectively, and there are 1000 moments on each 
solution path, so that a histogram of 40960 data can be drawn 
at each moment to obtain a plot of marginal distribution 
over time.
The purpose of these plots is to visualize and compare 
the different representations of the distributions 
under different SDEs.
All models were trained for 600K iterations with a 
batch size of 1024. $L$ in HOLD is set to 2.0,
$\xi$ is 6.0, and $\alpha$ is 0.04

% For CLD

% % model.sigma_min = 0.01
% %   model.sigma_max = 50
% %   model.num_scales = 1000
% %   model.beta_min = 0.1
% %   model.beta_max = 20.
% %   model.dropout = 0.1
% % m_inv=4.0, gamma=0.04, numerical_eps=1e-9, beta0=4.0, beta1=0.0

% For VPSDE

%training.batch_size = 1024
% beta0=4.0, beta1=0.0,

\subsection{Generate Data from 2D Multi-Swiss Rolls}
\label{appsec:gen_2d_swiss}

For experiments with 2D data, we 
use \texttt{sklearn.datasets.make\_swiss\_roll} 
with noise of 0.02 and multiplier fo 0.01 to generate 
the training data.
We assume that the target data has five swiss rolls, one is 
located at the 
origin, 
and the centers of the other four are at
(0.8, 0.8), (0.8, -0.8), (-0.8, -0.8), (-0.8, 0.8) respectively.
Use the same 5-layer MLP to predict scores, and train for 1600K 
iterations, with a batch size of 512.

\subsection{Generate High-Dimensional Data (Images)}
\label{appsec:gen_image}

We conducted experiments on two widely 
used image dataset benchmarks,  
CIFAR-10 and CelebA-HQ. The resolution 
of CIFAR-10 is 32$\times$32, with a total of 60,000 natural 
pictures, 50,000 as training and 10,000 as 
verification data. The original CelebA-HQ has 
30,000 high-definition 1024$\times$1024 celebrity face images. 
Considering our computing resources, this paper 
uses a 256$\times$256 version (CelebA-HQ-256).

\begin{figure}[th]%[th]%[htp]
    %\vspace{-0.4in}
    \centering
    %\begin{center}
    \hspace{-5mm}
    \includegraphics[width=1.0\linewidth]{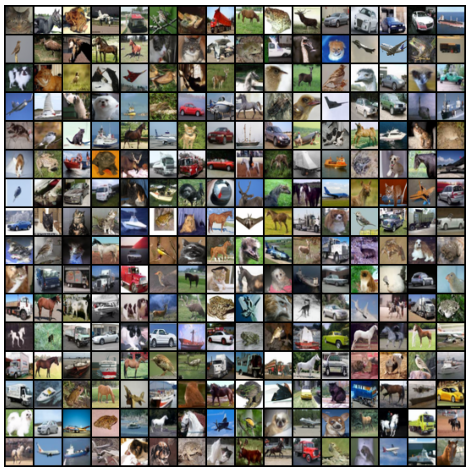}
    \hspace{-5mm}
    %\end{center}
    %\vspace{-2.5mm}
    \caption{
      Generated CIFAR-10 samples without cherry-picking.
    }
    \label{fig:256_cifar10_samples}
    %\vspace{-2.5mm}
    \end{figure}

The generative model on CIFAR-10 is trained on a 4-GPU server 
with 48G of memory per GPU. Figure~\ref{fig:256_cifar10_samples} shows 256 images 
generated by this model. It can be seen that the objects in these
generated images have various types, such as various animals, 
birds, natural scenery, cars, airplanes, ships, etc., and the 
image quality is also very high.

\begin{figure}[th]%[htp]
  %\vspace{-0.4in}
  \centering
  %\begin{center}
  \hspace{-5mm}
  \includegraphics[width=1.0\linewidth]{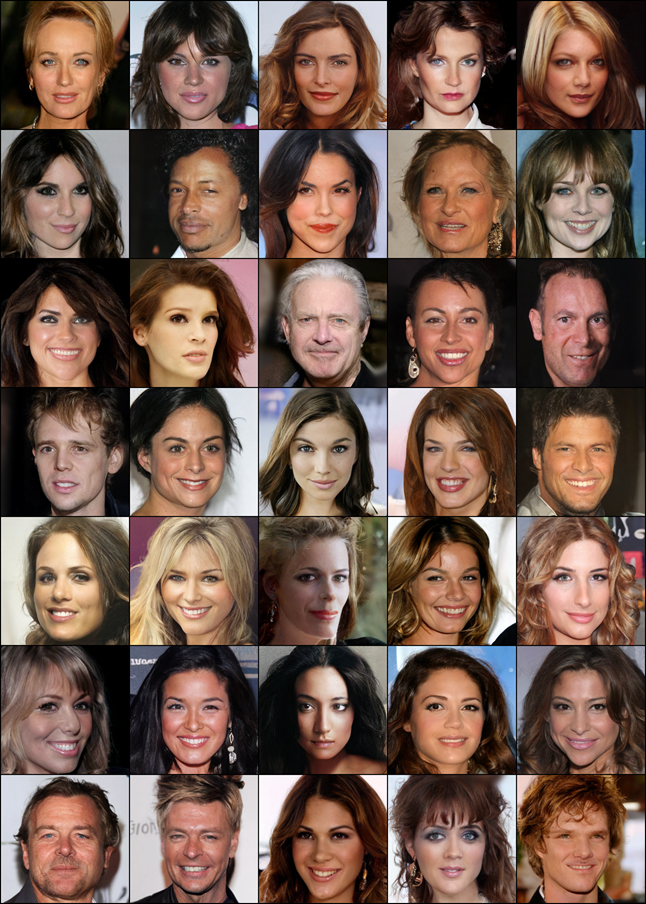}
  \hspace{-5mm}
  %\end{center}
  %\vspace{-2.5mm}
  \caption{
    Generated CelebA-HQ-256 samples.
  }
  \label{fig:selected_35_sample_38}
  %\vspace{-2.5mm}
  \end{figure}

\begin{figure}[th]%[htp]
  %\vspace{-0.4in}
  \centering
  %\begin{center}
  \hspace{-5mm}
  \includegraphics[width=1.0\linewidth]{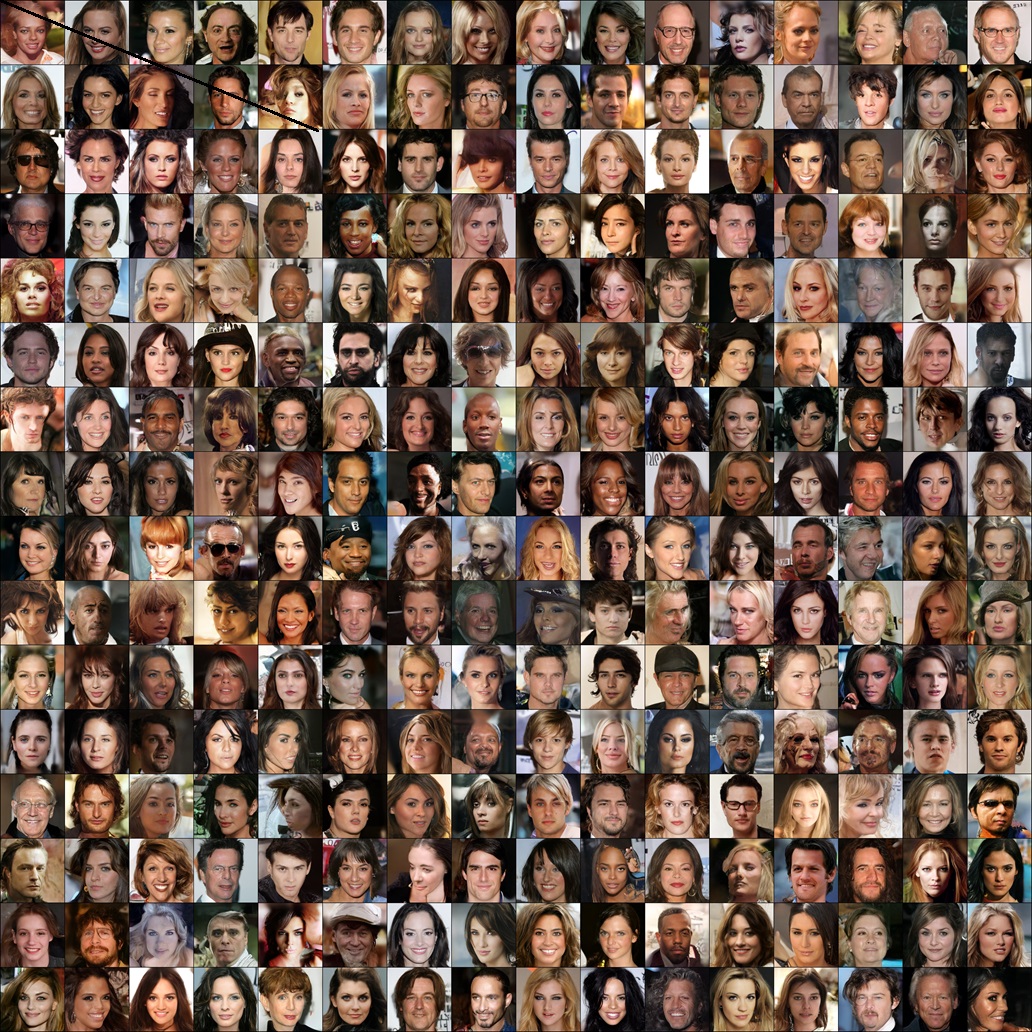}
  \hspace{-5mm}
  %\end{center}
  %\vspace{-2.5mm}
  \caption{
    Generated CelebA-HQ-256 samples without cherry-picking.
  }
  \label{fig:celeb_sample_6}
  %\vspace{-2.5mm}
  \end{figure}

When training the generative model on CelebA-HQ-256 face data, 
we trained it on an 8-GPU machine, each GPU has a memory of 24G. 
The batch size is 32 and the number of iterations is 2,400,000. 
Adam optimizer~\cite{kingma2014adam} is employed without 
weight decay. The initial learning 
rate is 2e-4 and the warmup is of 5000 iterations. For the neural 
network used for score prediction, we use the popular 
NCSP++~\cite{song2020improved}. 
There are 4 BigGAN~\cite{brock2018large} type residue blocks with 
 DDPM attention module~\cite{ho2020denoising}, whose resolution is 16. There are 
several generated faces in Figure~\ref{fig:selected_35_sample_38} 
and Figure~\ref{fig:celeb_sample_6}. It can 
be seen that the clarity of the faces is very high, and 
the coverage pattern is very complete. The generated faces 
have different genders, ages, skin colors, head postures, 
hair colors, expressions, different face action units and so on.

\begin{figure}[th]%[htp]
    %\vspace{-0.4in}
    \centering
    %\begin{center}
    \hspace{-5mm}
    \includegraphics[width=0.9\linewidth]{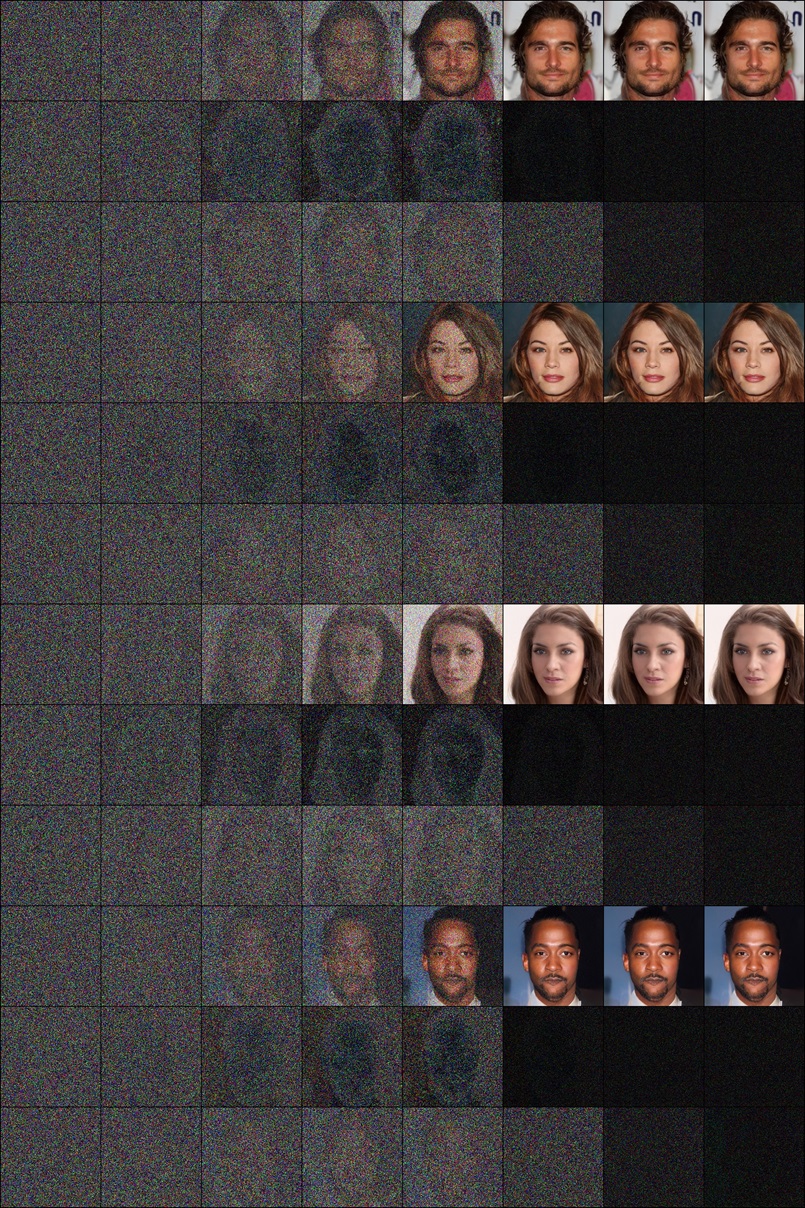}
    \hspace{-5mm}
    %\end{center}
    %\vspace{-2.5mm}
    \caption{
        When sampling by integration of time-reverse HOLD's ODE, 
        CelebA-HQ-256 images, velocities, 
        and accelerations are generated at different instants 
        (as a percentage of total steps are
        0., 0.5, 0.7, 0.8, 0.9, 0.99, 0.999, 0.99999) along 
        the solution path.
    }
    \label{fig:celeb_diff_qps_odeint}
    %\vspace{-2.5mm}
    \end{figure}

\begin{figure}[th]%[htp]
      %\vspace{-0.4in}
      \centering
      %\begin{center}
      \hspace{-5mm}
      \includegraphics[width=0.9\linewidth]{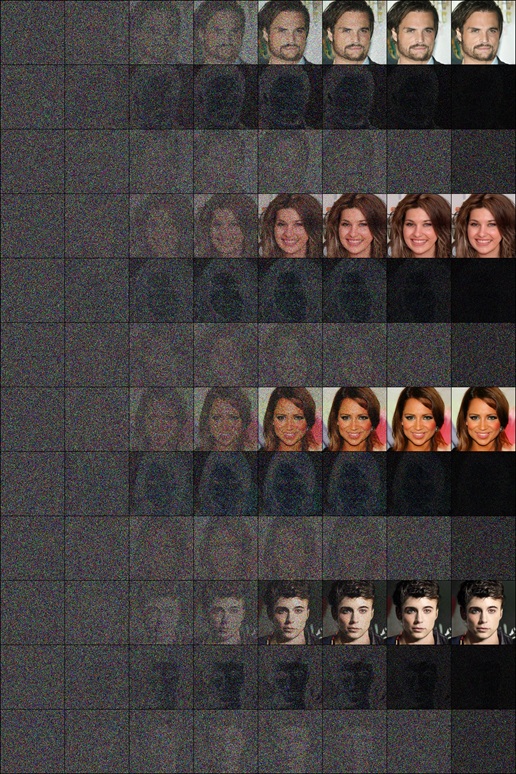}
      \hspace{-5mm}
      %\end{center}
      %\vspace{-2.5mm}
      \caption{
        When sampling with LT sampler, 
        CelebA-HQ-256 images, velocities, 
        and accelerations are generated at different instants 
        (as a percentage of total steps are
        0., 0.5, 0.7, 0.8, 0.9, 0.93, 0.97, 0.99) along 
        the solution path.
      }
      \label{fig:celeb_diff_qps_lt}
      %\vspace{-2.5mm}
      \end{figure}

Figure~\ref{fig:celeb_diff_qps_odeint} and 
Figure~\ref{fig:celeb_diff_qps_lt} show the images, 
velocities, and accelerations 
along the denoising path at different moments in the process of 
generating high-resolution faces. It can be seen how the speed 
and acceleration are evolved from the random state at the 
beginning, to 
the customized state formed according to the denoising of the 
generated faces in the middle, and to the final state with 
initial diffusion setting.

% \begin{figure}[th]%[htp]
%   %\vspace{-0.4in}
%   \centering
%   %\begin{center}
%   \hspace{-5mm}
%   \includegraphics[width=1.0\linewidth]{celeb_sample_6.png}
%   \hspace{-5mm}
%   %\end{center}
%   %\vspace{-2.5mm}
%   \caption{
%     Generated CelebA-HQ-256 samples.
%   }
%   \label{fig:celeb_sample_6}
%   %\vspace{-2.5mm}
%   \end{figure}

% You can have as much text here as you want. The main body must be at most $8$ pages long.
% For the final version, one more page can be added.
% If you want, you can use an appendix like this one, even using the one-column format.
%%%%%%%%%%%%%%%%%%%%%%%%%%%%%%%%%%%%%%%%%%%%%%%%%%%%%%%%%%%%%%%%%%%%%%%%%%%%%%%
%%%%%%%%%%%%%%%%%%%%%%%%%%%%%%%%%%%%%%%%%%%%%%%%%%%%%%%%%%%%%%%%%%%%%%%%%%%%%%%

\end{document}